\documentclass[sigconf]{acmart}

\usepackage{enumitem}
\usepackage{balance}

\usepackage{multirow}
\AtBeginDocument{%
  \providecommand\BibTeX{{%
    \normalfont B\kern-0.5em{\scshape i\kern-0.25em b}\kern-0.8em\TeX}}}

\copyrightyear{2023}
\acmYear{2023}
\setcopyright{acmlicensed}
\acmConference[MM '23]{Proceedings of the 31st ACM International Conference on Multimedia}{October 29-November 3, 2023}{Ottawa, ON, Canada}
\acmBooktitle{Proceedings of the 31st ACM International Conference on Multimedia (MM '23), October 29-November 3, 2023, Ottawa, ON, Canada}
\acmPrice{15.00}
\acmISBN{979-8-4007-0108-5/23/10}
\acmDOI{10.1145/3581783.3611807}

\acmSubmissionID{490}

\def\sota{state-of-the-art}
\def\reffig{Fig.}
\def\reftab{Table}

\def\refsec{Sec.}
\def\Nus{NuScenes}
\def\nus{nuScenes}
\def\dd{DDAD}
\def\ki{KITTI}
\def\fw{Diffusion-Augmented Depth Prediction}
\def\sxf{DADP}
\def\zs{noise predictor}
\def\loss{object-guided integrality loss}

\DeclareMathSymbol{@}{\mathord}{letters}{"3B}

\begin{document}
\title{Diffusion-Augmented Depth Prediction with Sparse Annotations}

\author{Jiaqi Li}
\email{lijiaqi_mail@hust.edu.cn}
\affiliation{%
  \institution{School of AIA, Huazhong University of Science and Technology}
  \country{}
}

\author{Yiran Wang}
\email{wangyiran@hust.edu.cn}
\affiliation{%
  \institution{School of AIA, Huazhong University of Science and Technology}
  \country{}
}
\authornote{Corresponding author.}

\author{Zihao Huang}
\email{zihaohuang@hust.edu.cn}
\affiliation{%
  \institution{School of AIA, Huazhong University of Science and Technology}
  \country{}
}

\author{Jinghong Zheng}
\email{deepzheng@hust.edu.cn}
\affiliation{%
  \institution{School of AIA, Huazhong University of Science and Technology}
  \country{}
}

\author{Ke Xian}
\email{ke.xian@ntu.edu.sg}
\affiliation{%
  \institution{S-Lab, Nanyang Technological University}
  \country{}
}

\author{Zhiguo Cao}
\email{zgcao@hust.edu.cn}
\affiliation{%
  \institution{School of AIA, Huazhong University of Science and Technology}
  \country{}
}

\author{Jianming Zhang}
\email{jianmzha@adobe.com}
\affiliation{
  \institution{Adobe Research}
  \country{}
}

\renewcommand{\shortauthors}{Jiaqi Li et al.}

\begin{abstract}
Depth estimation aims to predict dense depth maps. In autonomous driving scenes, sparsity of annotations makes the task challenging. Supervised models produce concave objects due to insufficient structural information. They overfit to valid pixels and fail to restore spatial structures. Self-supervised methods are proposed for the problem. Their robustness is limited by pose estimation, leading to erroneous results in natural scenes. In this paper, we propose a supervised framework termed \textbf{D}iffusion-\textbf{A}ugmented \textbf{D}epth \textbf{P}rediction \textbf{(DADP)}. We leverage the structural characteristics of diffusion model to enforce depth structures of depth models in a plug-and-play manner. An object-guided integrality loss is also proposed to further enhance regional structure integrality by fetching objective information. We evaluate DADP on three driving benchmarks and achieve significant improvements in depth structures and robustness. Our work provides a new perspective on depth estimation with sparse annotations in autonomous driving scenes.
\end{abstract}

\ccsdesc[500]{Computing methodologies~Scene understanding}

\keywords{depth prediction, diffusion model, autonomous driving}

\maketitle

\section{Introduction}
Depth prediction aims to predict dense depth maps, whether in supervised or self-supervised paradigm. However, sparse depth annotations in driving scenes pose an obstacle for the task. 

LiDAR is one primary acquisition equipment for driving datasets but only generates sparse annotations. \Nus{}~\cite{nus}, \dd{}~\cite{ddad}, and \ki{}~\cite{kitti} only have $0.24\%$, $1.85\%$, and $15.8\%$ pixels with valid ground truth respectively. 
In contrast, depth models are supposed to predict dense results with both accurate details and integral spatial structures. Due to insufficient structural information, supervised~\cite{kexian2020,midas,dpt} or self-supervised~\cite{monodepth2, surround,mcdp} methods produce failure predictions with concave objects, erroneous outcomes, or noticeable artifacts on autonomous driving scenarios~\cite{nus,ddad}.

\begin{figure}[!t]
    \centering
    \includegraphics[scale=0.85,trim=0 4 0 0,clip]{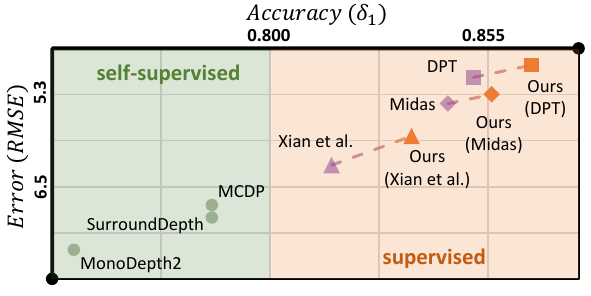}
    \caption{Comparisons with supervised~\cite{kexian2020,midas,dpt} or self-supervised~\cite{monodepth2,surround,mcdp} methods on \nus{}~\cite{nus}. The X-axis represents $\delta_1$. The Y-axis represents $\textbf{RMSE}$. Our \sxf{} achieves \sota{} performance and improvements with different depth predictors in a plug-and-play manner.}
    \label{fig:fig1}
\end{figure}

Supervised methods~\cite{kexian2020,midas,dpt,newcrfs,fmnet} employ various loss functions~\cite{silog,wacv19loss,kexian2020,midas,mai21,mai22report} to measure the discrepancy between output depth and ground truth. However, models fail to acquire sufficient structural information from sparse annotations of driving scenes. They overfit to pixels with valid ground truth and cannot preserve regional structures (\textit{i.e.}, complete shape of objects) and holistic structures. As shown in \reffig{}~\ref{fig:compare_sup}, supervised models~\cite{kexian2020,midas,dpt} trained on \nus{}~\cite{nus} produce concave areas on objects. They even exhibit holistically wrong spatial structures with striped artifacts, overfitting to horizontally-distributed annotations from LiDAR. As proved in \refsec{}~\ref{sec:motivation}, these defects cannot be solved
by common techniques for overfitting such as data augmentations or weight decay~\cite{weightdecay_adamw}. Previous supervised models and loss~\cite{dpt,midas,kexian2020} show limitations on sparse depth annotations of driving scenes.

To enhance spatial structures, recent works~\cite{monodepth2,packnet-sfm,fsm,surround,mcdp} explore self-supervised manner on driving scenes~\cite{nus,ddad}. SurroundDepth~\cite{surround} employs pseudo labels from Structure-from-Motion~\cite{colmapsfm} to pretrain their model. They utilize pose estimation and photometric loss~\cite{photometricloss} between six cameras to restore depth structures. MCDP~\cite{mcdp} conducts multi-camera prediction by projections between different views. However, self-supervised methods rely on pose estimation~\cite{monodepth2}, which is inaccurate in natural scenes and limits the robustness of those methods. As shown in \reffig{}~\ref{fig:compare_selfsup}, they produce erroneous results on night or rainy scenes. Besides, deploying multiple cameras on driving cars~\cite{surround} is inflexible and costly.

To overcome these challenges, we propose a novel supervised framework with sparse annotations termed \textbf{D}iffusion-\textbf{A}ugmented \textbf{D}epth \textbf{P}rediction \textbf{(\sxf{})}. Our method does not rely on pose estimation and multiple cameras, achieving better robustness than self-supervised methods~\cite{monodepth2,packnet-sfm,fsm,surround,mcdp}, especially for challenging night or rainy scenes. \sxf{} consists of a \zs{} and a depth predictor. The depth predictor can be different supervised single-image depth models. The core task is to enhance depth structures. Recent diffusion models in other tasks~\cite{stable,difseg_medical,segdiff21,difvideo} showcases favorable structural properties, which can span coherent parts of objects as shown in \reffig{}~\ref{fig:featurevis}. To acquire integral spatial structures, we introduce the \zs{} similar to diffusion models~\cite{ddpm,ddpmbeat} but in a plug-and-play manner. To be specific, we add Gaussian noise to input images. The \zs{} is trained to predict noise components. We fuse the structure-aware features from \zs{} and the detail-aware features from depth predictors, predicting depth maps with both accurate details and complete structures. The \zs{} can be adopted to off-the-shelf depth predictors in a plug-and-play manner. Besides, to further improve regional structure integrality of objects, we design our \loss{} that fetches objective structural information. 

Experiments are conducted on prevailing driving benchmarks \nus{}~\cite{nus}, \dd{}~\cite{ddad}, and KITTI~\cite{kitti}. \sxf{} effectively alleviates concave objects and artifacts produced by supervised depth predictors~\cite{kexian2020,midas,dpt}. Compared with self-supervised methods~\cite{monodepth2,mcdp,surround}, quantitative and qualitative results prove the robustness of our \sxf{} on challenging driving scenes with glare, reflections, rain, or weak-textured areas. As shown in \reffig{}~\ref{fig:fig1}, \sxf{} achieves \sota{} performance over previous supervised or self-supervised methods. We also adopt three different depth predictors~\cite{kexian2020,midas,dpt} and demonstrate the effectiveness of our plug-and-play paradigm. The main contributions can be summarized as follows:

\begin{itemize}[leftmargin=*]
    \item We present a plug-and-play framework with sparse depth annotations termed \textbf{D}iffusion-\textbf{A}ugmented \textbf{D}epth \textbf{P}rediction \textbf{(\sxf{})}.
    \item We propose a \zs{} to enforce depth structures in different depth predictors, utilizing structural properties of diffusion models to remedy sparse annotations in autonomous driving.
    \item We design \loss{} to further enhance the completeness of objects with objective structural information.
\end{itemize}

\section{Related Work}
\noindent\textbf{Depth Datasets for Autonomous Driving.}
Most driving depth datasets are captured by LiDAR and suffer from sparsity of ground truth. \Nus{}~\cite{nus} is a large-scale dataset captured by a sensor suite (with six cameras, one LiDAR, five RADAR, GPS, and IMU). It contains $1@000$ driving scenes in Boston and Singapore. \Nus{}~\cite{nus} is quite challenging for its low density of depth annotations ($0.24\%$) and its complex in-the-wild scenes.
\dd{}~\cite{ddad} is an urban driving dataset captured by six synchronized cameras and a high-resolution Luminar-H2 LiDAR. It is designed for long-range depth estimation in diverse urban scenes. The annotation density in \dd{}~\cite{ddad} is $1.85\%$. \ki{}~\cite{kitti} is another widely-used autonomous driving dataset with annotation density of $15.8\%$. It contains $61$ outdoor scenes with binocular views from a driving car.

\noindent\textbf{Supervised Depth Estimation.}
In recent years, supervised depth models~\cite{kexian2020,midas,adabins,nvds,mai21report} have significantly improved the depth accuracy. DPT~\cite{dpt} utilizes the vision transformer~\cite{VIT} for depth prediction and semantic segmentation. The structure-guided ranking loss is proposed by Xian \textit{et al.}~\cite{kexian2020} with a novel sampling strategy for learning from pseudo-depth data. Midas~\cite{midas} adopts a multi-objective learning strategy and trains depth models on mixing datasets. PackNet-SAN~\cite{sans} improve spatial structural integrity by jointly learning depth estimation and depth completion. 

Under sparse annotations of driving scenes, those supervised models produce incomplete objects and artifacts. In this work, we present our novel supervised framework \sxf{}. Thanks to the structure information from our \zs{} and the structure guidance from our \loss{}, the above-mentioned defects can be solved in a plug-and-play manner.

\noindent\textbf{Self-supervised Depth Estimation.}
To enhance depth structures on driving scenes~\cite{nus,ddad}, prior arts~\cite{packnet-sfm,surround,transdssl,fsm,mcdp} seek for the self-supervised paradigm. They jointly optimize a pose module and a depth module by photometric loss~\cite{photometricloss}.
Monodepth2~\cite{monodepth2} introduces the multi-scale sampling strategy and minimum re-projection loss to deal with obscuring situations. 
The cross-view transformer is proposed by SurroundDepth~\cite{surround} to improve self-supervised methods on surrounding views. MCDP~\cite{mcdp} iteratively optimizes depth results with neighboring cameras for consistent structures.

However, self-supervised methods~\cite{packnet-sfm,surround,transdssl,fsm,mcdp} rely on pose estimation, which is unreliable and significantly limits their robustness. In contrast, our \sxf{} achieves better robustness without pose estimation and multiple cameras.

\noindent\textbf{Diffusion Models.}
Sohl-Dickstein \textit{et al.}~\cite{ddpmini} first design the diffusion probabilistic model, learning to invert the diffusion process. Ho \textit{et al.}~\cite{ddpm} propose the DDPM framework to generate high-quality image samples. By learning the process of denoising, the diffusion model can generate images with both vivid details and reasonable structures. A series of improvements on model structures~\cite{improvedddpm,ddpmbeat} and sampling strategies~\cite{ddim,Analyticdpm,fastdif_iclr22} lead to the great success of diffusion models in generation tasks. They can synthesize realistic effects with reasonable spatial relations and structures according to prior conditions~\cite{ddpmbeat,classfreeddpm,palette,stable}. Those results indicate that diffusion models acquire a strong capability of structural representations. 

In our work, we leverage the structural information embedded in the diffusion models to enhance regional and holistic spatial structures for depth estimation on sparse driving scenes~\cite{nus,ddad,kitti}.

\begin{figure}[!h]
    \centering
    \includegraphics[scale=0.25,trim=0 79 0 0,clip]{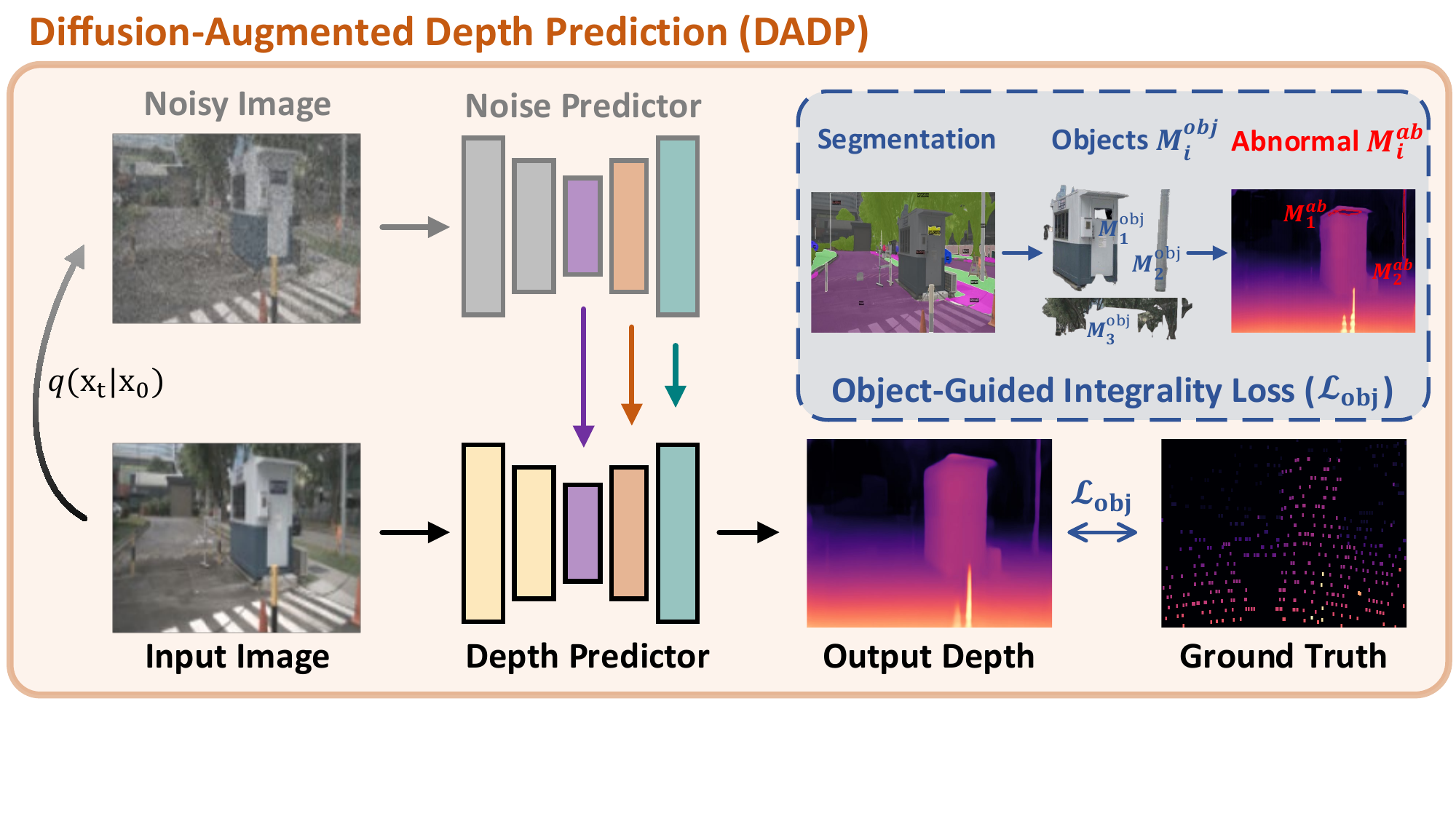}
    \caption{Overview of \sxf{} framework. \sxf{} contains a noise predictor and a depth predictor. The depth predictor can be different single-image depth models. The noise predictor is an Unet~\cite{ddpmbeat} to predict the noise component of diffusion step $q(x_t|x_0)$. We fuse structure-aware features from \zs{} and detail-aware features of depth predictor to enhance depth structures with sparse annotations on driving scenes. We also propose \loss{} to improve objective structural integrity. Our loss guides abnormal regions $\mathbf{M}^{ab}_{i}$ with incorrect depth variation to normal depth range in different objects. Better viewed when zoomed in.} 
    \label{fig:pipeline}
\end{figure}

\section{Proposed Method}

\subsection{Overview}
\reffig{}~\ref{fig:pipeline} showcases the technical pipeline of our \fw{} (\sxf{}) framework. On autonomous driving scenarios only with sparse annotations~\cite{nus,ddad}, the key problem is to enforce regional and holistic spatial structures without the unreliable pose estimation in prior arts~\cite{surround,monodepth2,mcdp,fsm,transdssl}. \sxf{} contains two main components: a depth predictor and a \zs{}. The depth predictor can be different single-image depth models such as Xian \textit{et al.}~\cite{kexian2020}, Midas~\cite{midas}, and DPT~\cite{dpt}. The \zs{} is a Unet~\cite{improvedddpm,ddpmbeat} to predict noise components of the diffusion process. The \zs{} and diffusion models~\cite{ddpm} are trained to predict noise components and generate noise-free images from the noisy ones. In this way, the \zs{} acquires favorable structural properties as demonstrated in \refsec{}~\ref{sec:motivation}. We adopt the \zs{} to restore depth structures. Specifically, we fuse the structure-aware features from the \zs{} and the detail-aware features from the depth predictor by feature fusion modules (FFM)~\cite{FFM1,FFM2}. Ultimately, depth predictors can restore both accurate details and integral spatial structures.

We conduct a two-stage training procedure. First, we train the \zs{} on RGB images with the subtask of unconditional image generation as DDPM~\cite{ddpm,ddpmbeat,improvedddpm}. Once the \zs{} is trained, its parameters are fixed. It can be directly adopted to the training of different depth predictors in a plug-and-play manner. The next step is the supervised training for depth predictors. To further improve the regional structural integrality, we design our object-guided integrality loss which fetches the spatial structural information of different objects. To be specific, we utilize a \sota{} panoptic segmentation model~\cite{mask2former} to segment different objects. We focus on abnormal regions within each object, \textit{i.e.}, concave areas with incorrect depth variation. Our loss guides those abnormal regions to integral depth structures.

During inference, along with the \zs{}, depth predictors predict depth maps with regionally and holistically complete spatial structures, removing the incomplete objects, concave areas, and artifacts produced by previous supervised depth predictors~\cite{kexian2020,midas,dpt} on sparse driving scenes~\cite{nus,ddad}. Without relying on pose estimation, our \sxf{} achieves better robustness than self-supervised frameworks~\cite{surround,monodepth2,mcdp} on challenging scenarios.

\subsection{Noise Predictor}
\label{sec:noiseprediction}
\noindent \textbf{Preliminaries.}
Diffusion models~\cite{ddpm,ddpmbeat,Difsamantic,ddim} are latent variable models for generative tasks, which are trained to denoise Gaussian-blurred images and reverse the diffusion process. If we denote a noise-free input RGB image as $x_0\in\mathbb{R}^{h\times w \times 3}$ and the maximum diffusion step as $T$, the diffusion process of diffusion step $t\in\{0,1,\cdots,T\}$ can be formulated as follows:
\begin{equation}
    \begin{gathered}
    q\left(x_t \mid x_0\right):=\mathcal{N}\left(x_t ; \sqrt{\bar{\alpha}_t} x_0,\left(1-\bar{\alpha}_t\right) I\right)\,,\\ 
    x_t=\sqrt{\bar{\alpha}_t} x_0+\sqrt{1-\bar{\alpha}_t} \epsilon, \quad \epsilon \sim \mathcal{N}(0,1)\,, \label{equ:diffusionprocess}
    \end{gathered}
\end{equation}
where ${\alpha}_t:=1-\beta_t$ and $\bar{\alpha}_t:=\prod_{s=0}^t \alpha_s$. $\beta$ is the noise variance schedule as DDPM~\cite{ddpm}. $x_t\in\mathbb{R}^{h\times w \times 3}$ represents the resulting high-noise image of diffusion step $t$. $\mathcal{N}$ denotes Gaussian noise.

As for the denoising process, a \zs{} $\epsilon_\theta(x_t,t)$ is trained to reverse the diffusion process and restore the noise-free $x_0$. Each denoising step is approximated by a Gaussian distribution:
\begin{equation}
    \label{equ:(dif3)}
    p_\theta\left(x_{t-1} \mid x_t\right):=\mathcal{N}\left(x_{t-1} ; \mu_\theta\left(x_t, t\right), \Sigma_\theta\left(x_t, t\right)\right)\;.
\end{equation}

The \zs{} $\epsilon_\theta(x_t,t)$ predicts the noise component of step $t$ and obtains the mean value $\mu_\theta(x_t,t)$ of above-mentioned Gaussian distribution by linear combination of $x_t$.
The covariance predictor $\Sigma_\theta\left(x_t, t\right)$ can be fixed~\cite{ddpm} or learned~\cite{ddpmbeat}. By iterating the denoising process, the noise-free image $x_0$ can be recovered.

\noindent \textbf{Depth Structure Augmentation.}
We adopt an Unet as the \zs{} with multi-resolution attention~\cite{ddpmbeat} and BigGAN up-/down-sampling~\cite{biggan}. It is trained on RGB images with mean square error (MSE) between the predicted and actual noise components. In \refsec{}~\ref{sec:motivation}, we prove that the \zs{} acquires a strong ability of structural representations by denoising RGB images. After the training of \zs{}, we freeze its parameters and utilize its structure-aware features to augment depth structures in a plug-and-play manner. See supplementary for details of the \zs{}.

As shown in \reffig{}~\ref{fig:pipeline}, we fuse the structure-aware features from \zs{} and the detail-aware features from depth predictor by feature fusion modules (FFM)~\cite{FFM1,FFM2}. The CNN decoder of the depth predictor gradually improves spatial resolutions and predicts depth results. Specifically, if we denote the block indexes of the \zs{} as $b$, we leverage structure-aware features $\{t=50,b=12\}$, $\{t=100,b=8\}$, and $\{t=150,b=5,6,7\}$ from the \zs{}. Resolutions of the structure-aware features are adjusted to $1/2$, $1/4$, and $1/8$ of input resolution respectively. 

The ultimate step is to train depth predictors~\cite{kexian2020,midas,dpt}. Directly training depth predictors on highly sparse driving scenes~\cite{nus,ddad} will produce incomplete objects, concave areas, and artifacts by previous supervised training procedures and loss functions~\cite{silog,midas,kexian2020}. With the structure information from \zs{}, our \sxf{} can remove those defects and predict depth maps with regionally and holistically integral spatial structures. The \zs{} can be adapted to different depth predictors~\cite{kexian2020,midas,dpt} in a plug-and-play manner. As for the supervision between predicted depth and ground truth, we adopt the commonly-applied affinity invariant loss~\cite{midas}. More importantly, to further improve the regional integrality of objects, we propose to use the objective structural information as guidance, which will be illustrated in the next section.

\subsection{Object-Guided Integrality Loss}
\label{sec:methodloss}
With the structural information from the \zs{} described above, our \sxf{} is capable to restore spatial structures from sparse depth annotations of autonomous driving scenarios. To further improve the integrity of different objects, we propose the \loss{} as regional structural guidance.

The design of our \loss{} is based on a simple prior that the depth values inside a certain object should be continuous and smooth. Our loss focuses on the areas with abnormal depth variation inside each object and guides them back to normal depth values. Thus, we need to extract each object region from the input image. In our implementation, we utilize the \sota{} Mask2Former~\cite{mask2former} to perform panoptic segmentation and extract the object mask of each instance. Here, we denote the object masks as $\mathbf{M}^{obj}_{i}, i \in \{1,2,\cdots ,K\}$, where $K$ represents the number of objects in the input image.

Specifically, given the noise-free input image $\mathbf{x_0}$, we denote the predicted and ground truth depth as $\hat{\mathbf{d}}$ and $\hat{\mathbf{d}}^{*}$, which are aligned to zero translation and unit scale as MiDaS~\cite{midas} to deal with varied scale and shift of different depth predictors~\cite{kexian2020,dpt,midas}.  
For each object $\mathbf{M}^{obj}_i$, we segment the corresponding region of predicted depth, denoted by $\hat{\mathbf{d}}_{i}$. The operation can be easily achieved by dot product between object mask $\mathbf{M}^{obj}_i$ and predicted depth $\hat{\mathbf{d}}$. Similarly, We obtain the list $\hat{\mathbf{d}}^{*}_{i}$ of valid ground truth values within the object. 

The next step is to find the areas with abnormal depth variation inside each object. We consider the normal depth range in object $\mathbf{M}^{obj}_i$ with upper bound $\mathbf{U}_{i}$ and lower bound $\mathbf{L}_{i}$ as follows:
\begin{equation}
    \begin{gathered}
    \mathbf{U}_{i}=(1+\alpha)* \textit{max} 
   (\textit{max}(\hat{\mathbf{d}}^{*}_{i}),\textit{median}(\hat{\mathbf{d}}_{i}))\;, \\
    \mathbf{L}_{i}=(1-\alpha)* \textit{min}(\textit{min}(\hat{\mathbf{d}}^{*}_{i}),\textit{median}(\hat{\mathbf{d}}_{i}))\;, 
    \label{equ:loss1}
    \end{gathered}
\end{equation}
where $\alpha$ is the tolerance factor. To be mentioned, for some small objects with no valid ground truth, \textit{i.e.}, $\hat{\mathbf{d}}^{*}_{i}$ is empty, we use the median value of $\hat{\mathbf{d}}_{i}$ with tolerance factor $\alpha$ for $\mathbf{U}_{i}$ and $\mathbf{L}_{i}$. In contrast, when pixels with valid ground truth exist in the object, the maximum and minimum value of $\hat{\mathbf{d}}^{*}_{i}$ will be adopted. 

Our loss encourages depth values inside the object to be continuous and smooth. Pixels with depth values out of the normal depth range will be divided into abnormal areas.
However, the segmentation model~\cite{mask2former} cannot handle some complex occlusions, \textit{e.g.}, sky regions with tree branches. Pixels that actually do not belong to the certain object should be removed from $\mathbf{M}^{obj}_i$. For this problem, we conduct k-means clustering~\cite{kmeans} on the input image $\mathbf{x_0}$ within the object $\mathbf{M}^{obj}_i$. Pixels with the lowest $20\%$ cosine similarity to their corresponding clustering centers are removed. If we denote those pixels to be removed as the mask $\mathbf{M}^{occ}_i$, the final abnormal region $\mathbf{M}_i^{ab}$ as shown in \reffig{}~\ref{fig:pipeline} is presented as follows:
\begin{equation}
    \label{equ:loss2}
    \mathbf{M}_i^{ab}= (\hat{\mathbf{d}}_{i}>\mathbf{U}_{i}) \cup (\hat{\mathbf{d}}_{i}<\mathbf{L}_{i}) -\mathbf{M}^{occ}_i\;.
\end{equation}

Finally, the abnormal depth values should be guided to the normal depth range. If we denote the pixels within $\mathbf{M}_i^{ab}$ as $p$, our \loss{} can be formulated as:
\begin{equation}
    \label{equ:loss3}
    \mathcal{L}_{obj}\left(\hat{\mathbf{d}} ,\mathbf{U},\mathbf{L}\right)=\frac{1}{K} \sum_{i=1}^{K}  \sum_{p=1}^{N_i}  \textit{min}\left(\left|\hat{ \mathbf{d}}_{i}^{(p)}-\mathbf{U}_{i} \right|,\left|\hat{\mathbf{d}}_{i}^{(p)}-\mathbf{L}_{i}\right| \right)\;,
\end{equation}
where $N_i$ is the number of pixels in $\mathbf{M}_i^{ab}$. See supplementary for more visual results of the masks in our loss.

\subsection{Implementation Details}
\label{sec:impldetail}
\noindent \textbf{Noise Predictor.} For our \zs{}, the maximum diffusion step $T$ is set to $1000$. we resize images to $256\times256$ and adopt a batch size of $8$ for training. We use the Adam optimizer with learning rate $1e-4$.
 
The mean square error (MSE) between the predicted and actual noise components is used for supervision. The parameters of our \zs{} are fixed after the first stage of training.

\noindent \textbf{Depth Predictor.} To prove our plug-and-play manner, we adopt Xian \textit{et al.}~\cite{kexian2020}, DPT-hybrid~\cite{dpt} and Midas-v2~\cite{midas} as depth predictors in our experiments. Xian \textit{et al.}~\cite{kexian2020} is trained from scratch, while DPT~\cite{dpt} and Midas~\cite{midas} are fine-tuned on \nus{}~\cite{nus} and \dd{}~\cite{ddad} datasets from their pretrained checkpoints.

For the training of depth predictors, following prior arts~\cite{surround,sans,mcdp}, we utilize the training resolution of $640\times352$ on \nus{} dataset~\cite{nus} and $640\times384$ on \dd{} dataset~\cite{ddad}. We train depth predictors for $5$ epochs on the \nus{}~\cite{nus} and $20$ epochs on \dd{}~\cite{ddad} with a batch size of $16$. The initial learning rate is set to $5e-5$ and decreases by $1e-5$ for every five epochs. For KITTI~\cite{kitti} dataset, we follow the same standard training procedure as BTS~\cite{bts}. Except for our \loss{}, we adopt the commonly-applied affinity invariant loss~\cite{midas,dpt} $\mathcal{L}_{af}$ between predicted depth and ground truth. See supplementary for more details. The overall loss $\mathcal{L}$ can be expressed as:
\begin{equation}
    \label{equ:(loss5)}
    \mathcal{L}=\mathcal{L}_{af}+\lambda \mathcal{L}_{obj}\;,
\end{equation}
where the coefficient $\lambda$ is $0.1$. We set the tolerance factor $\alpha = 0.1$.

\section{Experiments}
In this section, we evaluate our \fw{} (\sxf{}) framework on prevailing autonomous driving datasets \nus{}~\cite{nus}, \dd{}~\cite{ddad}, and \ki{}~\cite{kitti}. We first briefly describe the evaluation protocol and datasets in \refsec{}~\ref{sec:datasets}. Some experimental results are shown to further expound on our motivations in \refsec{}~\ref{sec:motivation}. The quantitative and qualitative comparisons with \sota{} approaches are shown in \refsec{}~\ref{sec:compare}. We also conduct ablation studies and prove the effectiveness of our design in \refsec{}~\ref{sec:ab}.

\begin{figure}[!t]
    \centering
    \includegraphics[scale=0.35,trim=0 0 0 0,clip]{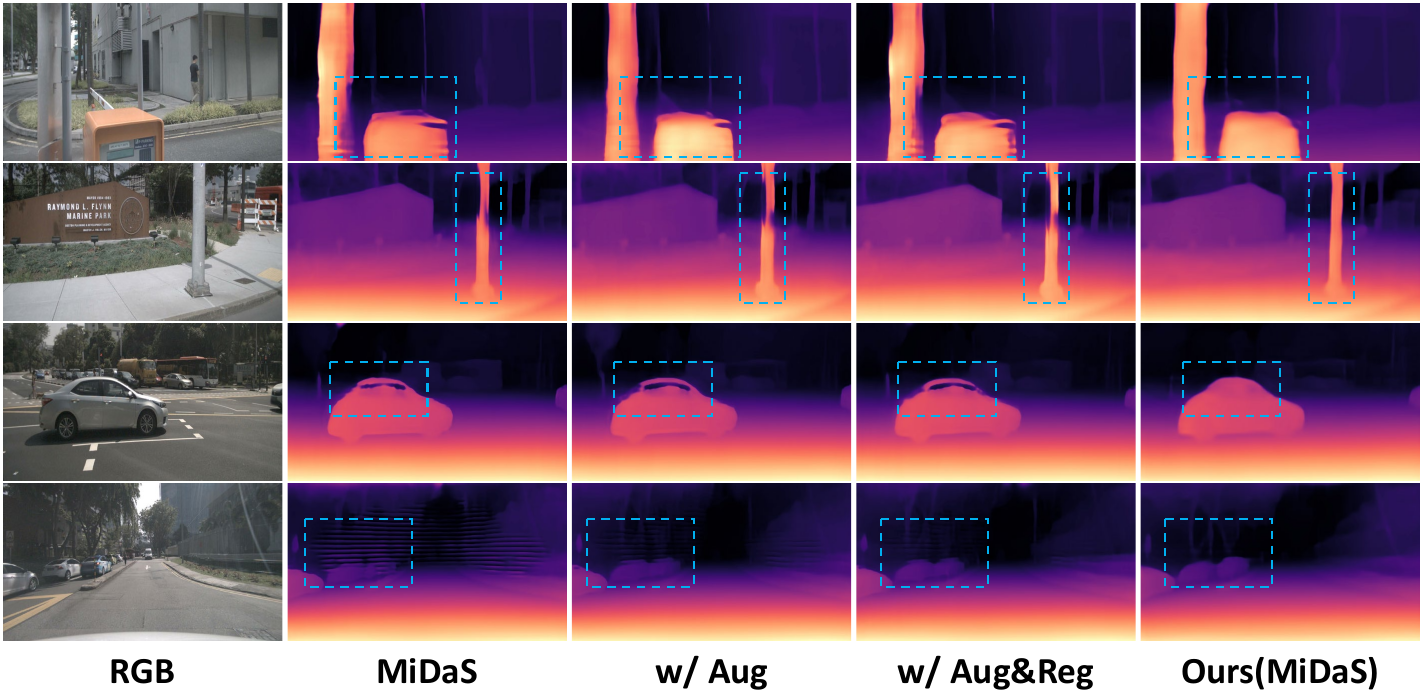}
    \caption{Visual results with data augmentations and weight decay~\cite{weightdecay_adamw}. We train Midas~\cite{midas} on \nus{} dataset~\cite{nus} with several data augmentations (Aug) and $L_2$ regularization (Reg) for weight decay. We highlight regions with prominent difference in dashed rectangular. The incomplete objects, concave areas, and artifacts cannot be solved by those common techniques for overfitting. Better viewed when zoomed in.}
    \label{fig:compare_crop}
\end{figure}

\subsection{Evaluation Protocol and Datasets}
\label{sec:datasets}

\noindent \textbf{Evaluation Protocol.}
For supervised depth models, we compare our \sxf{} with Xian \textit{et al.}~\cite{kexian2020}, Midas~\cite{midas} , and DPT~\cite{dpt} as different depth predictors, which demonstrates our effective plug-and-play manner. Those models~\cite{kexian2020,midas,dpt} are trained on \nus{}~\cite{nus} and \dd{}~\cite{ddad} with their original training pipeline. On \dd{}~\cite{ddad} dataset, we also compare our approach with previous \sota{} supervised framework PackNet-SAN~\cite{sans}. Following PackNet-SAN~\cite{sans}, we evaluate supervised methods~\cite{kexian2020,midas,dpt,sans} on four views of \dd{}~\cite{ddad}, due to self-occlusion in the other two views.

Self-supervised methods~\cite{surround,monodepth2,packnet-sfm,fsm,mcdp} with pose estimation of multiple views are prevailing on autonomous driving datasets~\cite{nus,ddad}. Consequently, we also compare our \sxf{} with them, even though the quantitative comparisons between supervised and self-supervised methods might be unfair. We mainly compare and demonstrate the robustness of our \sxf{} especially in challenging driving scenes with glare and reflections at night, rainy scenes, or weak-textured areas. Our approach does not utilize the unreliable camera poses. Following previous \sota{} self-supervised MCDP~\cite{mcdp} on \dd{}~\cite{ddad} dataset, we compare with self-supervised approaches~\cite{kexian2020,midas,dpt,sans} on six views of \dd{}~\cite{ddad}.

\noindent \textbf{Datasets.}
We mainly compare those supervised or self-supervised methods on \nus{}~\cite{nus} and \dd{}~\cite{ddad} datasets. To demonstrate the robustness of our framework, we further evaluate different approaches with daytime and nighttime scenes on \nus{}~\cite{nus} dataset. For sufficient comparisons and evaluations, we also compare with those methods on \ki{}~\cite{kitti} dataset. See supplementary for the experimental results on \ki{}~\cite{kitti} dataset.

\noindent \textbf{Evaluation Metrics.}
We evaluate the performance of different approaches with the commonly-applied depth metrics including $Abs\,Rel$, $Sq\,Rel$, $RMSE$, and $\delta_i\,(i=1,2,3)$.

\begin{table*}
    \begin{center}
    \resizebox{\textwidth}{!}{
    \begin{tabular}{lccccccccccccc}
    \toprule
    \multirow{2}{*}{Method} &
    \multicolumn{6}{c}{\Nus{}} &&
    \multicolumn{6}{c}{\dd{}} \\
    \cmidrule{2-7} \cmidrule{9-14}
        & Abs Rel$\downarrow$ & Sq Rel$\downarrow$ & RMSE$\downarrow$ & $\delta_1\uparrow$ & $\delta_2\uparrow$ & $\delta_3\uparrow$ & &
        Abs Rel$\downarrow$ & Sq Rel$\downarrow$ & RMSE$\downarrow$ & $\delta_1\uparrow$   & $\delta_2\uparrow$ & $\delta_3\uparrow$ \\
    \midrule
    Xian \textit{et al.}~\cite{kexian2020}& $0.147$ & $1.375$ & $6.266$ & $0.799$ & $0.913$ & $0.957$ && $0.150$ & $3.281$ & $12.784$& $0.810$ & $0.914$ & $0.954$\\
    PackNet~\cite{packnet}             & $-$ & $-$ & $-$ & $-$ & $-$ & $-$ && $0.125$ & $2.158$ & $11.245$ & $0.836$ & $0.929$ & $0.962$\\
    MiDaS~\cite{midas}          & $0.122$ & $1.106$ & $5.485$ & $0.844$ & $0.933$ & $0.964$ && $0.137$ & $2.904$ & $11.870$& $0.836$ & $0.929$ & $0.964$\\
    DPT~\cite{dpt}              & $0.121$ & $1.074$ & $\underline{5.315}$ & $0.851$ & $0.934$ & $0.965$ && $0.134$ & $2.630$ & $11.913$& $0.837$ & $0.930$ & $0.964$\\
    PackNet-SAN~\cite{sans}         & $-$ & $-$ & $-$ & $-$ & $-$ & $-$ && $\underline{0.119}$ & $\textbf{1.931}$ & $10.852$ & $0.850$ & $0.936$ & $\textbf{0.977}$\\
    \midrule
    Ours(Xian \textit{et al.})  & $0.131$ & $1.141$ & $5.952$ & $0.824$ & $0.924$ & $0.963$ && 
    $0.138$ & $2.297$ & $11.150$& $0.838$ & $0.934$ & $0.967$\\
    Ours(MiDaS)                 & $\underline{0.117}$ & $\underline{1.084}$ & $5.370$ & $\underline{0.856}$ & $\underline{0.938}$ & $\underline{0.967}$ && 
    $0.122$ & $2.227$ & $\underline{10.406}$& $\underline{0.866}$ & $\underline{0.945}$ & $0.970$\\
    Ours(DPT)                   & $\textbf{0.112}$ & $\textbf{1.010}$ & $\textbf{5.181}$ & $\textbf{0.867}$ & $\textbf{0.941}$ & $\textbf{0.968}$ && 
    $\textbf{0.118}$ & $\underline{2.140}$ & $\textbf{10.130}$ & $\textbf{0.870}$ & $\textbf{0.946}$ & $\underline{0.972}$  \\
    \bottomrule
    \end{tabular}
    }
\end{center}

\caption{Comparisons with supervised depth estimation approaches on \nus{}~\cite{nus} and \dd{}~\cite{ddad}. The first five rows are results of previous \sota{} supervised models. We show our results with different depth predictors in the last three rows, which demonstrate the effectiveness of our plug-and-play manner. Following PackNet-SAN~\cite{sans}, we evaluate supervised approaches on four views of \dd{}~\cite{ddad}. Best performance is in boldface. Second best is underlined.}
\label{tab:supervised_big}
\vspace{-20pt}
\end{table*}

\begin{table*}
    \begin{center}
    \resizebox{\textwidth}{!}{
    \begin{tabular}{lccccccccccccc}
    \toprule
    \multirow{2}{*}{Method} &
    \multicolumn{6}{c}{\Nus{}} &&
    \multicolumn{6}{c}{\dd{}} \\
    \cmidrule{2-7} \cmidrule{9-14}
        & Abs Rel$\downarrow$ & Sq Rel$\downarrow$ & RMSE$\downarrow$ & $\delta_1\uparrow$ & $\delta_2\uparrow$ & $\delta_3\uparrow$ & &
        Abs Rel$\downarrow$ & Sq Rel$\downarrow$ & RMSE$\downarrow$ & $\delta_1\uparrow$   & $\delta_2\uparrow$ & $\delta_3\uparrow$ \\
    \midrule
    Monodepth2~\cite{monodepth2} (-M)  & $0.287$ & $3.349$ & $7.184$ & $0.641$ & $0.845$ & $0.925$ && $0.217$ & $3.641$ & $12.962$& $0.699$ & $0.877$ & $0.939$\\
    PackNet-SfM~\cite{packnet-sfm} (-M)     & $0.309$ & $2.891$ & $7.994$ & $0.547$ & $0.796$ & $0.899$ && $0.234$ & $3.802$ & $13.253$& $0.672$ & $0.860$ & $0.931$\\
    FSM~\cite{fsm}                 & $0.334$ & $2.845$ & $7.786$ & $0.508$ & $0.761$ & $0.894$ && $0.229$ & $4.589$ & $13.520$& $0.677$ & $0.867$ & $0.936$\\
    TransDSSL~\cite{transdssl}       & $-$ & $-$ & $-$ & $-$ & $-$ & $-$ && $0.151$ & $3.591$ & $14.350$& $-$ & $-$ & $-$\\
    SurroundDepth~\cite{surround}    & $0.245$ & $3.067$ & $6.835$ & $0.719$ & $0.878$ & $0.935$ && $0.200$ & $3.392$ & $12.270$& $0.740$ & $0.894$ & $0.947$\\
    MCDP~\cite{mcdp}                 & $0.237$ & $3.030$ & $6.822$ & $0.719$ & $-$   & $-$       && $0.193$ & $3.111$ & $12.264$& $0.811$ & $-$     & $-$\\
    \midrule
    Ours(Xian \textit{et al.})       & $0.131$ & $1.141$ & $5.952$ & $0.824$ & $0.924$ & $0.963$ &&        
    $0.146$ & $2.352$ & $10.778$ & $0.826$ & $0.927$ & $0.962$\\
    Ours(MiDaS)                 & $\underline{0.117}$ & $\underline{1.084}$ & $\underline{5.370}$ & $\underline{0.856}$ & $\underline{0.938}$ & $\underline{0.967}$ && 
    $\underline{0.132}$ & $\textbf{2.312}$ & $\underline{10.087}$ & $\underline{0.856}$ & $\underline{0.938}$ & $\underline{0.966}$\\
    Ours(DPT)                        & $\textbf{0.112}$ & $\textbf{1.010}$ & $\textbf{5.181}$ & $\textbf{0.867}$ & $\textbf{0.941}$ & $\textbf{0.968}$ && 
    $\textbf{0.130}$ & $\underline{2.338}$ & $\textbf{9.994}$ & $\textbf{0.860}$ & $\textbf{0.939}$ & $\textbf{0.966}$  \\
    \bottomrule
    \end{tabular}
    }
\end{center}

\caption{Comparisons with self-supervised depth estimation approaches on \nus{}~\cite{nus} and \dd{}~\cite{ddad} datasets. The first six rows are results of previous \sota{} self-supervised models. Our results with different depth predictors are in the last three rows. Best results are highlighted in bold. Second best is underlined. $(-M)$ indicates occlusion masking~\cite{surround} in \dd{}~\cite{ddad}.}
\label{tab:self-supervised big}
\end{table*}

\begin{figure}[!t]
    \centering
    \includegraphics[scale=0.30,trim=0 0 0 0,clip]{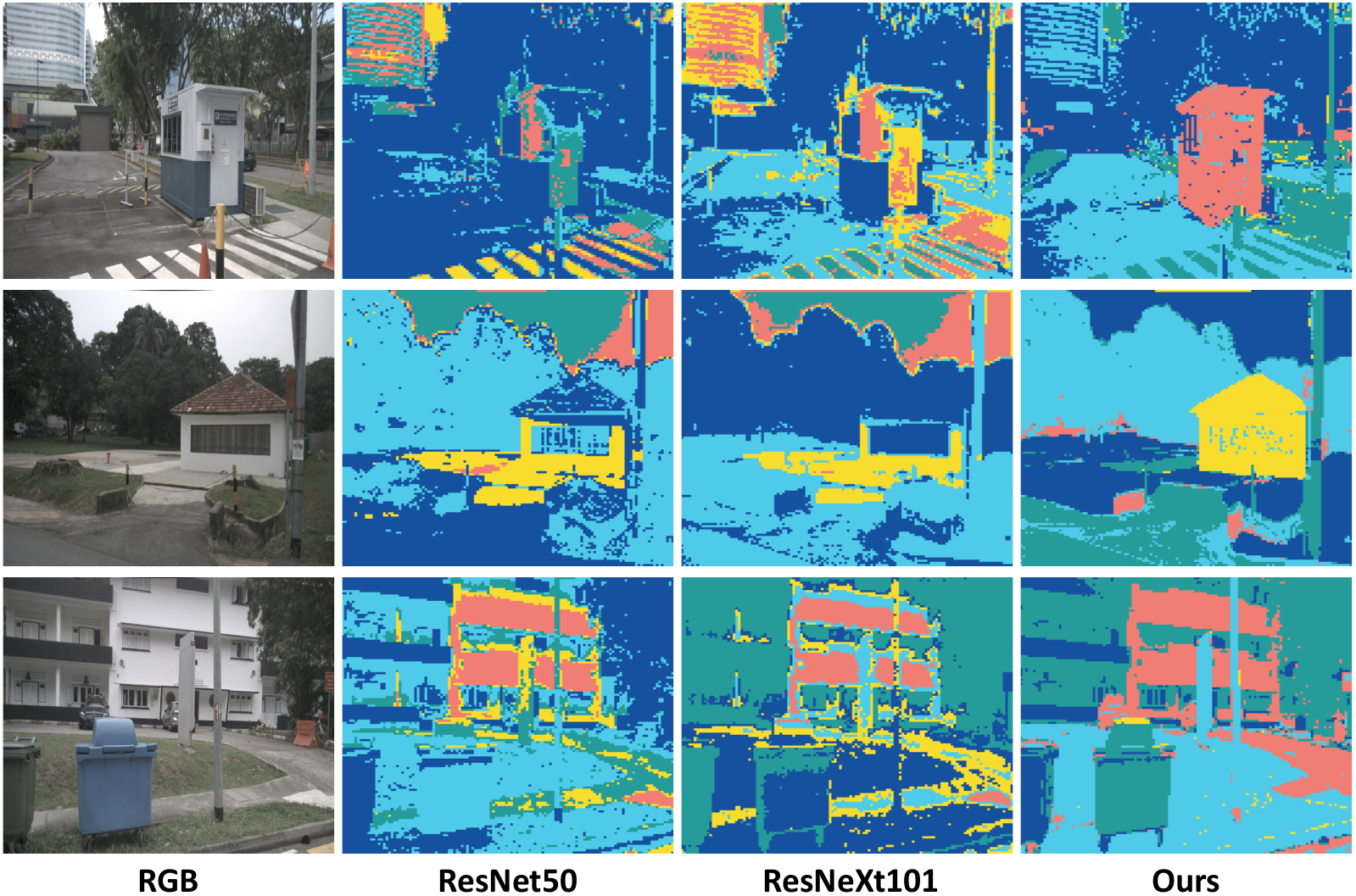}
    \caption{Visual comparisons of features. We visualize feature maps from CNN encoders~\cite{resnet,resnext} and the structure-aware features in our \zs{} by k-means~\cite{kmeans} clustering (k=5). Our \zs{} shows better representations of spatial structures, which can span coherent parts of objects.}
    \label{fig:featurevis}

\end{figure}
\subsection{Motivation Review}
\label{sec:motivation}
Dominant autonomous driving depth datasets \Nus{}~\cite{nus} and \dd{}~\cite{ddad} only have $0.24\%$ and $1.85\%$ pixels with depth ground truth. Directly training fully supervised depth estimation models~\cite{kexian2020,midas,dpt} on those highly sparse driving scenes produces concave regions, incomplete objects, or even some artifacts in predicted depth maps. Due to the lack of structural information in the sparse annotations, models could overfit to pixels with valid ground truth and fail to establish regional and holistic spatial structures. 

Experimental results prove that those defects cannot be solved by common techniques for overfitting such as data augmentations or weight decay~\cite{weightdecay_adamw}. We train Midas~\cite{midas} with weight decay~\cite{weightdecay_adamw} and several data augmentations on \nus{}~\cite{nus} dataset. Specifically, we adopt random cropping, horizontal flipping, and color jittering as data augmentations (Aug). We also apply $L_2$ regularization (Reg) for weight decay. As shown in \reffig{}~\ref{fig:compare_crop}, the above-mentioned defects still remain with data augmentations and weight decay~\cite{weightdecay_adamw}. The incomplete objects and concave areas cannot be improved without sufficient structural information and guidance.

Previous supervised learning approaches~\cite{kexian2020,midas,dpt,sans} cannot handle the sparsity problem for autonomous driving. They do not conduct experiments on \nus{}~\cite{nus} dataset while only PackNet-SAN~\cite{sans} shows results on \dd{}~\cite{ddad} dataset. To establish complete spatial structures on those sparse driving scenes~\cite{nus,ddad}, most previous works~\cite{surround,mcdp,fsm,monodepth2,packnet-sfm} seek for the self-supervised learning manner with pose estimation in multiple views. However, camera poses are inaccurate and unreliable on many challenging scenes and limit the robustness of those self-supervised approaches~\cite{surround,mcdp,fsm,monodepth2,packnet-sfm}. As shown in \reffig{}~\ref{fig:compare_selfsup}, previous Monodepth2~\cite{monodepth2} and SurroundDepth~\cite{surround} produce erroneous predictions on night or rainy scenes.   

We are devoted to solving this dilemma. We prefer not to rely on the pose estimation and self-supervised manner considering their limited robustness on natural scenes. The key problem is to enforce integral spatial structures under the condition of highly sparse depth annotations. An intuitive idea is to fuse spatial features extracted from RGB images by convolutional encoders~\cite{resnet,resnext}. Specifically, we replace the \zs{} with widely-used CNN encoders~\cite{resnet,resnext} and fuse the extracted features into the depth predictor. However, concave objects and artifacts are not settled. More quantitative and visual depth comparisons can be found in our supplementary. We visualize the feature maps of the CNN encoders~\cite{resnet,resnext} by k-means clustering~\cite{kmeans}. As shown in \reffig{}~\ref{fig:featurevis}, detailed RGB information rather than spatial structures are presented  by CNNs~\cite{resnet,resnext}. For better spatial structures, motivated by recent diffusion models~\cite{scoredif,DifferentiableDepth,ddpmbeat,Difsamantic,ddim}, we propose our \sxf{} with the \zs{}. The \zs{} is trained to denoise and generate noise-free images from noisy ones. This subtask can enforce better representations of spatial structures in the structure-aware features as shown in \reffig{}~\ref{fig:featurevis}. With the structural information from the \zs{}, our \sxf{} predicts depth results with integral regional and holistic structures, significantly alleviating the incomplete objects and artifacts. 

Besides, to further guide integral regional structures of objects, we design the \loss{}. Our loss focuses on the areas with abnormal depth variation inside a certain object and guides the depth values to normal depth range. Overall, our \sxf{} and \loss{} significantly improve depth structures on sparse autonomous driving scenarios~\cite{nus,ddad} and obtain better robustness without using pose estimation of multiple views. 

\begin{figure*}[!h]
    \centering
    \includegraphics[scale=0.55,trim=0 0 0 0,clip]{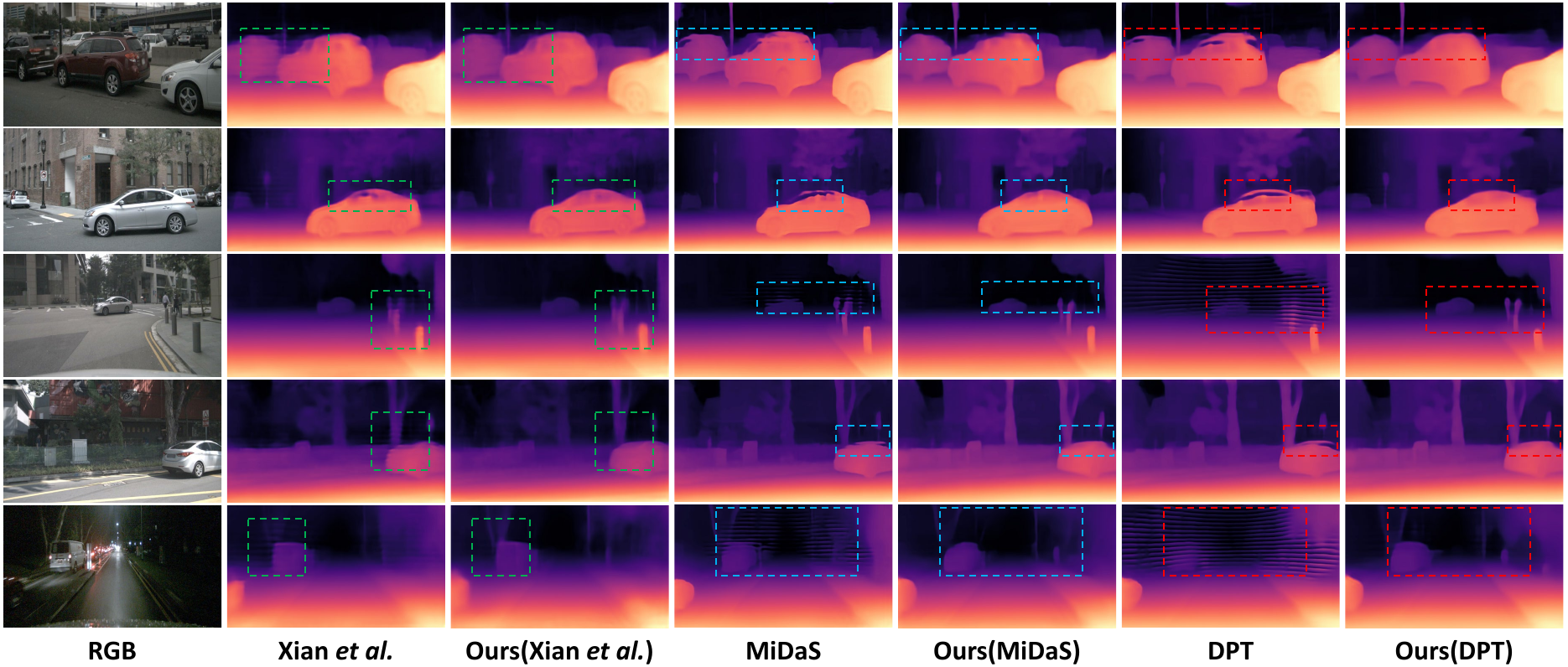}
    \caption{Visual comparisons with supervised depth prediction methods on \nus{} dataset~\cite{nus}. We adopt supervised models~\cite{kexian2020,midas,dpt} as different depth predictors in our \sxf{} framework. Our approach effectively releases concave areas, incomplete objects, and artifacts produced by depth predictors. Regions with prominent difference are highlighted in dashed rectangular.}
    \label{fig:compare_sup}
\end{figure*}

\subsection{Comparisons with state-of-the-art results}
\label{sec:compare}

\noindent \textbf{Comparisons with supervised methods.} 
We compare our \sxf{} with previous supervised depth estimation approaches~\cite{packnet,sans,kexian2020,midas,dpt}. We present quantitative comparisons in \reftab{}~\ref{tab:supervised_big} and qualitative results in \reffig{}~\ref{fig:compare_sup}. For PackNet~\cite{packnet} and PackNet-SAN~\cite{sans}, we report their official depth metrics on \dd{} dataset~\cite{ddad}. For Xian~\textit{et al.}~\cite{kexian2020}, Midas~\cite{midas}, and DPT~\cite{dpt}, we train their models on \nus{} and \dd{} datasets~\cite{nus,ddad} with their original training pipeline. In \reffig{}~\ref{fig:compare_sup}, we can observe that previous supervised frameworks~\cite{kexian2020,midas,dpt} produce obvious concave areas, incomplete objects, or even artifacts. With the structural information from the \zs{}, our \sxf{} effectively releases those defects and achieves \sota{} performance as shown in \reftab{}~\ref{tab:supervised_big}. 

Besides, we also fit the three different supervised depth models~\cite{kexian2020,midas,dpt} as depth predictors into our \sxf{} framework. Our approach can effectively deal with different depth predictors in a plug-and-play manner. With Xian~\textit{et al.}~\cite{kexian2020}, Midas~\cite{midas}, and DPT~\cite{dpt} as depth predictors on \dd{}~\cite{ddad} dataset, our \sxf{} shows $2.8\%$, $3.0\%$, and $3.3\%$ improvements of $\delta_1$ respectively. The quantitative metrics cannot fully reflect our improvements due to the sparse ground truth. Both the depth metrics and visualizations demonstrate the effectiveness of our plug-and-play manner.

\begin{figure}[!t]
    \centering
    \includegraphics[scale=0.45,trim=0 0 0 0,clip]{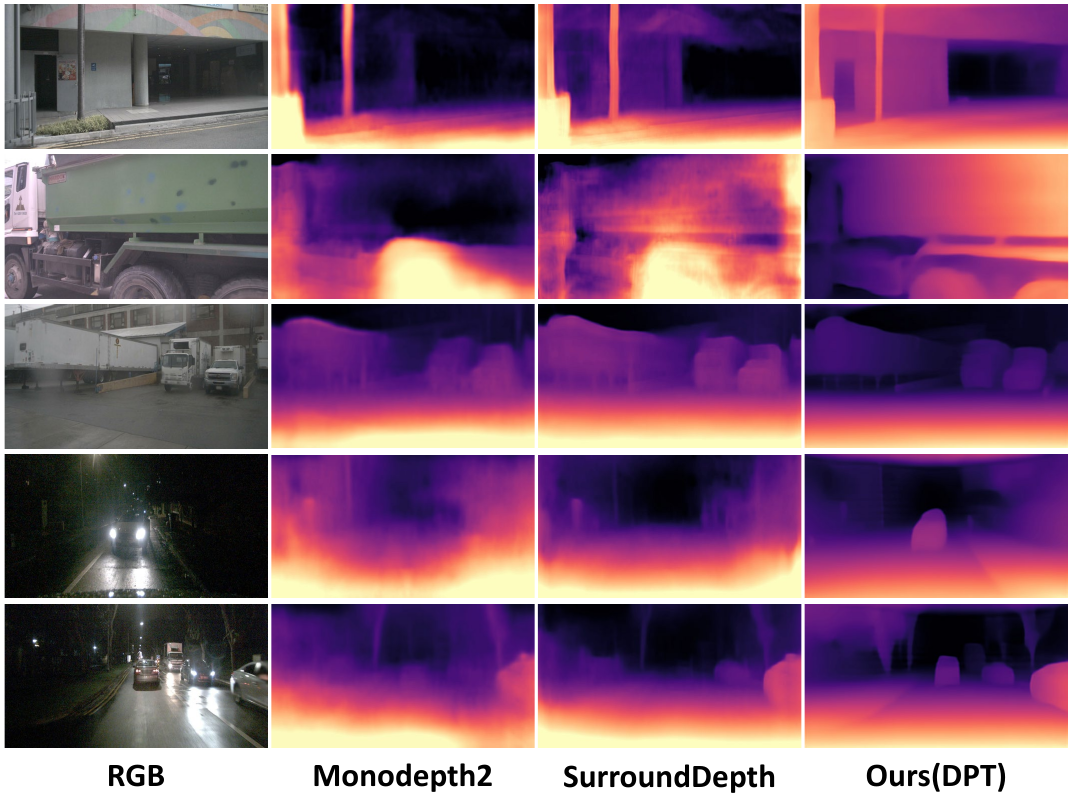}
    \caption{Visual comparisons with self-supervised methods~\cite{monodepth2,surround}. Without relying on pose estimation, our \sxf{} shows significantly better robustness in challenging scenes with glare and reflections at night, rainy scenes, or weak-textured areas. Better viewed when zoomed in.}
    \label{fig:compare_selfsup}
\end{figure}

\begin{table}
  \caption{Comparisons on daytime and nighttime scenes of \nus{} dataset~\cite{nus}. We evaluate both supervised and self-supervised approaches~\cite{monodepth2,surround,midas,dpt}. Our approach shows better robustness on both daytime and nighttime scenarios.}
  \setlength{\tabcolsep}{2pt}
  \label{tab:nusc night}
  \resizebox{\columnwidth}{!}{
  \begin{tabular}{lcccccc} 
    \toprule
    Method & Abs Rel$\downarrow$ & Sq Rel$\downarrow$ & RMSE$\downarrow$ & $\delta_1$$\uparrow$ & $\delta_2$$\uparrow$ & $\delta_3$$\uparrow$ \\
    \midrule
    \multicolumn{7}{c}{Daytime Scenes} \\
    \midrule
    Monodepth2~\cite{monodepth2}   & $0.263$ & $2.772$ & $7.061$ & $0.660$ & $0.854$ & $0.928$\\
    SurroundDepth ~\cite{surround}   & $0.236$ & $3.002$ & $6.754$ & $0.738$ & $0.889$ & $0.940$\\
    MiDaS~\cite{midas}          & $0.118$ & $1.176$ & $5.212$ & $0.859$ & $0.937$ & $0.967$\\
    DPT~\cite{dpt}              & $0.115$ & $1.049$ & $5.269$ & $0.863$ & $0.938$ & $0.967$\\
    Ours(MiDaS)                 & $0.111$ & $1.049$ & $5.292$ & $0.871$ & $0.943$ & $0.969$\\
    Ours(DPT)                   & $\textbf{0.106}$ & $\textbf{0.984}$ & $\textbf{5.120}$ & $\textbf{0.879}$ & $\textbf{0.946}$ & $\textbf{0.970}$\\
    \midrule
    \multicolumn{7}{c}{Nighttime Scenes} \\
    \midrule
    Monodepth2~\cite{monodepth2}   & $0.585$ & $18.715$ & $12.119$ & $0.484$ & $0.732$ & $0.851$\\
    SurroundDepth ~\cite{surround}   & $0.330$ & $3.638$ & $7.563$ & $0.542$ & $0.784$ & $0.890$\\
    MiDaS~\cite{midas}          & $0.198$ & $1.699$ & $6.197$ & $0.697$ & $0.871$ & $0.940$\\
    DPT~\cite{dpt}              & $0.177$ & $1.297$ & $\textbf{5.727}$ & $0.743$ & $0.891$ & $0.948$\\
    Ours(MiDaS)                 & $0.182$ & $1.397$ & $6.068$ & $0.722$ & $0.884$ & $0.948$\\
    Ours(DPT)                   & $\textbf{0.166}$ & $\textbf{1.243}$ & $5.731$ & $\textbf{0.757}$ & $\textbf{0.896}$ & $\textbf{0.952}$\\
    \bottomrule
\end{tabular}
}

\end{table}

\noindent \textbf{Comparisons with self-supervised frameworks.}
Due to the challenging sparse annotations for autonomous driving, most previous depth estimation methods on \nus{}~\cite{nus} and \dd{}~\cite{ddad} datasets are in the self-supervised paradigm with pose estimation. We compare our \sxf{} with previous self-supervised frameworks~\cite{monodepth2,packnet-sfm,fsm,surround,mcdp,transdssl} in \reftab{}~\ref{tab:self-supervised big} and \reffig{}~\ref{fig:compare_selfsup}. We know that it might be unfair to compare supervised approaches with self-supervised methods. However, it is meaningful to explore supervised learning frameworks on sparse driving scenes. The robustness of self-supervised methods is limited due to the unreliable camera poses. Accurate pose estimation itself is a challenging task especially on in-the-wild images with glare and reflections at night, rainy scenes, or weak-textured areas. Whether compared with supervised or self-supervised methods, our \sxf{} shows better qualitative and quantitative performance. Compared with self-supervised approaches, our \sxf{} does not utilize camera poses and achieves better robustness on challenging driving scenes as shown in \reffig{}~\ref{fig:compare_selfsup}.

\noindent \textbf{Robustness on Challenging Driving Scenes.}
 To further demonstrate our robustness on challenging driving scenes, we evaluate previous supervised and self-supervised approaches~\cite{monodepth2,surround,midas,dpt} with daytime and nighttime scenes respectively on \nus{} dataset~\cite{nus}. The results are shown in \reftab{}~\ref{tab:nusc night}. Compared with \sota{} self-supervised SurroundDepth~\cite{surround} with pose estimation, our \sxf{} showcases $21\%$ $\delta_1$ improvements in the challenging nighttime scenes. Compared with the supervised models, our method also outperforms DPT~\cite{dpt} by $4.2\%$ in $Sq\,Rel$ for nighttime. Whether compared with previous supervised or self-supervised approaches, our framework shows better robustness on challenging autonomous driving scenes with highly sparse depth annotations. 

\begin{figure}[!t]
    \centering
    \includegraphics[scale=0.47,trim=0 0 0 0,clip]{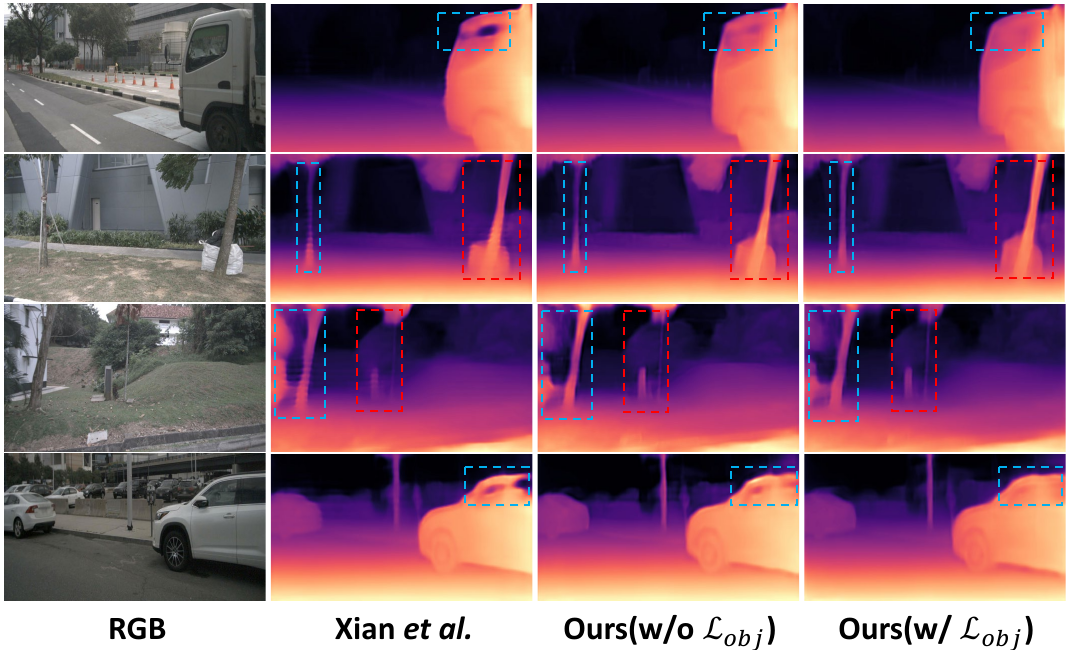}
    \caption{Qualitative ablation of \sxf{} framework. The \zs{} enforces integral structures and removes stripped artifacts in the depth predictor~\cite{kexian2020}. The \loss{} further improves regional structural completeness of objects. Better viewed when zoomed in.}
    \label{fig:compare_abnetwork}
\end{figure}
\begin{table}
  \setlength{\tabcolsep}{2.5pt}
  \caption{Ablation of our \zs{} and \loss{} with Xian \textit{et al.}~\cite{kexian2020} as the depth predictor.}
  \label{tab:ab1}
  \resizebox{\columnwidth}{!}{
  \begin{tabular}{lcccccc} 
    \toprule
    Method & Abs Rel$\downarrow$ & Sq Rel$\downarrow$ & RMSE$\downarrow$ & $\delta_1$$\uparrow$ & $\delta_2$$\uparrow$ & $\delta_3$$\uparrow$ \\
    \midrule
    Xian \textit{et al.}~\cite{kexian2020}     & $0.183$ & $1.375$ & $6.266$ & $0.799$ & $0.913$ & $0.957$\\
    Ours ($w/o \mathcal{L}_{obj}$)       & $0.140$ & $1.292$ & $6.001$ & $0.820$ & $0.921$ & $0.960$\\
    Ours ($w/ \mathcal{L}_{obj}$)       & $\textbf{0.132}$ & $\textbf{1.141}$ & $\textbf{5.951}$ & $\textbf{0.824}$ & $\textbf{0.925}$ & $\textbf{0.963}$\\
    \bottomrule
\end{tabular}
}
\vspace{-5pt}
\end{table}

\subsection{Ablation Studies}
\label{sec:ab}
\noindent \textbf{Effectiveness of \sxf{} Framework.}
We ablate the \zs{} and our \loss{} to demonstrate the effectiveness of our design. In this experiment, we adopt Xian \textit{et al.}~\cite{kexian2020} as the depth predictor on \nus{} dataset~\cite{nus}. Quantitative results are shown in \reftab{}~\ref{tab:ab1}. Thanks to the structural information from the \zs{}, our \sxf{} achieves $2.1\%$ improvement on $\delta_1$ compared with the depth predictor. By adding our \loss{} for supervision, the depth accuracy further improves. To be mentioned, the quantitative metrics with sparse ground truth cannot fully reflect our improvements in depth structures.

Visual comparisons are shown in \reffig{}~\ref{fig:compare_abnetwork}. The \zs{} enforces spatial structures and removes the stripped artifacts in the depth predictor~\cite{kexian2020}. Meanwhile, our \loss{} can further improve the regional object completeness, for example, removing the concave areas on the window of the car. These structural improvements could not be presented by the depth metrics. Those concave areas might have few pixels with valid ground truth.

\begin{figure}[!t]
    \centering
    \includegraphics[scale=1.0,trim=15 0 0 15,clip]{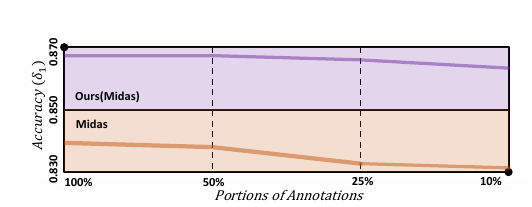} 
    
    \caption{Ablation of annotation density on \dd{} dataset. \dd{}~\cite{ddad} dataset has $\textbf{1.85\%}$ pixels with valid ground truth. We train our \sxf{} with Midas~\cite{midas} as the depth predictor using different portions of the valid pixels.  Our \sxf{} framework shows strong robustness on high annotation sparsity.}
    \label{fig:compare_bfbgt} 
    
\end{figure}

\begin{table}
   \setlength{\tabcolsep}{2.5pt}  

  \caption{Ablation of diffusion steps on \nus{}~\cite{nus} dataset.}
  
  \label{tab:ab_step}
  \resizebox{\columnwidth}{!}{
  \begin{tabular}{lcccccc} 
    \toprule
    Diffusion Steps & Abs Rel$\downarrow$ & Sq Rel$\downarrow$ & RMSE$\downarrow$ & $\delta_1$$\uparrow$ & $\delta_2$$\uparrow$ & $\delta_3$$\uparrow$ \\
    \midrule
    $(0,50,100)$     & $0.135$ & $1.207$ & $5.964$ & $0.820$ & $0.923$ & $0.962$\\
    $(100,150,200)$  & $0.137$ & $1.214$ & $6.010$ & $0.817$ & $0.922$ & $0.962$\\
    $(50,100,150)$   & $\textbf{0.132}$ & $\textbf{1.141}$ & $\textbf{5.951}$ & $\textbf{0.824}$ & $\textbf{0.925}$ & $\textbf{0.963}$\\
    \bottomrule
\end{tabular}
}

\end{table}

\noindent \textbf{Density of Depth Annotations.}
Different autonomous driving datasets~\cite{nus,kitti,ddad} possess different annotation density. To demonstrate the robustness of our framework on high sparsity, we randomly sample several portions of valid pixels for training on \dd{} dataset~\cite{ddad}. As shown in \reffig{}~\ref{fig:compare_bfbgt}, with lower annotation density, our \sxf{} shows stronger robustness than the depth predictor~\cite{midas}. When the portion decreases to the extremely low $10\%$, \textit{i.e.}, only with $0.185\%$ pixels with valid annotations, our \sxf{} still showcases $3.2\%$ improvements on $\delta_1$  over Midas~\cite{midas} as the depth predictor. 

\noindent \textbf{Diffusion Steps.}
In this experiment, we ablate different diffusion steps on \nus{}~\cite{nus} dataset. The experimental results are shown in \reftab{}~\ref{tab:ab_step}. With larger diffusion steps, features become more abstract and structural while remaining less details. Structure-aware features of the \zs{} should achieve a balance of structural and detailed information. As a consequence, we choose the best-performing diffusion steps of $(50,100,150)$ in our experiments.

\section{Conclusion}
Previous supervised depth methods lead to concave areas and artifacts on sparse driving scenes. In this paper, we propose a supervised framework termed \fw{} (\sxf{}).  With the structural information from our \zs{} and structural guidance from our \loss{}, \sxf{} effectively alleviates those defects in a plug-and-play manner. Compared with the prevailing self-supervised methods, our \sxf{} is more robust on challenging scenes without relying on pose estimation. Further analysis shows the efficacy of our design and proves that \sxf{} can adapt to varied annotation density. Our
work shows strong effectiveness in the field of autonomous driving.

\section{ACKNOWLEDGMENTS}
This work was funded by the National Natural Science Foundation of China under Grant No. U1913602 and was supported by Adobe. 

\vfill\eject
\bibliographystyle{ACM-Reference-Format}
\balance
\bibliography{main}


\begin{thebibliography}{57}


\ifx \showCODEN    \undefined \def \showCODEN     #1{\unskip}     \fi
\ifx \showDOI      \undefined \def \showDOI       #1{#1}\fi
\ifx \showISBNx    \undefined \def \showISBNx     #1{\unskip}     \fi
\ifx \showISBNxiii \undefined \def \showISBNxiii  #1{\unskip}     \fi
\ifx \showISSN     \undefined \def \showISSN      #1{\unskip}     \fi
\ifx \showLCCN     \undefined \def \showLCCN      #1{\unskip}     \fi
\ifx \shownote     \undefined \def \shownote      #1{#1}          \fi
\ifx \showarticletitle \undefined \def \showarticletitle #1{#1}   \fi
\ifx \showURL      \undefined \def \showURL       {\relax}        \fi
\providecommand\bibfield[2]{#2}
\providecommand\bibinfo[2]{#2}
\providecommand\natexlab[1]{#1}
\providecommand\showeprint[2][]{arXiv:#2}

\bibitem[Amit et~al\mbox{.}(2021)]%
        {segdiff21}
\bibfield{author}{\bibinfo{person}{Tomer Amit}, \bibinfo{person}{Eliya
  Nachmani}, \bibinfo{person}{Tal Shaharbany}, {and} \bibinfo{person}{Lior
  Wolf}.} \bibinfo{year}{2021}\natexlab{}.
\newblock \showarticletitle{Segdiff: Image segmentation with diffusion
  probabilistic models}.
\newblock \bibinfo{journal}{\emph{arXiv preprint arXiv:2112.00390}}
  (\bibinfo{year}{2021}).
\newblock


\bibitem[Bao et~al\mbox{.}(2022)]%
        {Analyticdpm}
\bibfield{author}{\bibinfo{person}{Fan Bao}, \bibinfo{person}{Chongxuan Li},
  \bibinfo{person}{Jun Zhu}, {and} \bibinfo{person}{Bo Zhang}.}
  \bibinfo{year}{2022}\natexlab{}.
\newblock \showarticletitle{Analytic-dpm: an analytic estimate of the optimal
  reverse variance in diffusion probabilistic models}.
\newblock  (\bibinfo{year}{2022}).
\newblock


\bibitem[Baranchuk et~al\mbox{.}(2022)]%
        {Difsamantic}
\bibfield{author}{\bibinfo{person}{Dmitry Baranchuk}, \bibinfo{person}{Ivan
  Rubachev}, \bibinfo{person}{Andrey Voynov}, \bibinfo{person}{Valentin
  Khrulkov}, {and} \bibinfo{person}{Artem Babenko}.}
  \bibinfo{year}{2022}\natexlab{}.
\newblock \showarticletitle{Label-Efficient Semantic Segmentation with
  Diffusion Models}. In \bibinfo{booktitle}{\emph{International Conference on
  Learning Representations}}.
\newblock


\bibitem[Bhat et~al\mbox{.}(2021)]%
        {adabins}
\bibfield{author}{\bibinfo{person}{Shariq~Farooq Bhat},
  \bibinfo{person}{Ibraheem Alhashim}, {and} \bibinfo{person}{Peter Wonka}.}
  \bibinfo{year}{2021}\natexlab{}.
\newblock \showarticletitle{Adabins: Depth estimation using adaptive bins}. In
  \bibinfo{booktitle}{\emph{Proceedings of the IEEE/CVF Conference on Computer
  Vision and Pattern Recognition (CVPR)}}. \bibinfo{pages}{4009--4018}.
\newblock


\bibitem[Brock et~al\mbox{.}(2019)]%
        {biggan}
\bibfield{author}{\bibinfo{person}{Andrew Brock}, \bibinfo{person}{Jeff
  Donahue}, {and} \bibinfo{person}{Karen Simonyan}.}
  \bibinfo{year}{2019}\natexlab{}.
\newblock \showarticletitle{Large scale GAN training for high fidelity natural
  image synthesis}.
\newblock  (\bibinfo{year}{2019}).
\newblock


\bibitem[Caesar et~al\mbox{.}(2020)]%
        {nus}
\bibfield{author}{\bibinfo{person}{Holger Caesar}, \bibinfo{person}{Varun
  Bankiti}, \bibinfo{person}{Alex~H. Lang}, \bibinfo{person}{Sourabh Vora},
  \bibinfo{person}{Venice~Erin Liong}, \bibinfo{person}{Qiang Xu},
  \bibinfo{person}{Anush Krishnan}, \bibinfo{person}{Yu Pan},
  \bibinfo{person}{Giancarlo Baldan}, {and} \bibinfo{person}{Oscar Beijbom}.}
  \bibinfo{year}{2020}\natexlab{}.
\newblock \showarticletitle{nuScenes: A Multimodal Dataset for Autonomous
  Driving}. In \bibinfo{booktitle}{\emph{Proceedings of the IEEE/CVF Conference
  on Computer Vision and Pattern Recognition (CVPR)}}.
  \bibinfo{pages}{11618--11628}.
\newblock


\bibitem[Cheng et~al\mbox{.}(2022)]%
        {mask2former}
\bibfield{author}{\bibinfo{person}{Bowen Cheng}, \bibinfo{person}{Ishan Misra},
  \bibinfo{person}{Alexander~G Schwing}, \bibinfo{person}{Alexander Kirillov},
  {and} \bibinfo{person}{Rohit Girdhar}.} \bibinfo{year}{2022}\natexlab{}.
\newblock \showarticletitle{Masked-attention mask transformer for universal
  image segmentation}. In \bibinfo{booktitle}{\emph{Proceedings of the IEEE/CVF
  Conference on Computer Vision and Pattern Recognition (CVPR)}}.
  \bibinfo{pages}{1290--1299}.
\newblock


\bibitem[Dhariwal and Nichol(2021)]%
        {ddpmbeat}
\bibfield{author}{\bibinfo{person}{Prafulla Dhariwal} {and}
  \bibinfo{person}{Alexander Nichol}.} \bibinfo{year}{2021}\natexlab{}.
\newblock \showarticletitle{Diffusion models beat gans on image synthesis}.
\newblock \bibinfo{journal}{\emph{Advances in Neural Information Processing
  Systems}}  \bibinfo{volume}{34} (\bibinfo{year}{2021}),
  \bibinfo{pages}{8780--8794}.
\newblock


\bibitem[Dosovitskiy et~al\mbox{.}(2020)]%
        {VIT}
\bibfield{author}{\bibinfo{person}{Alexey Dosovitskiy}, \bibinfo{person}{Lucas
  Beyer}, \bibinfo{person}{Alexander Kolesnikov}, \bibinfo{person}{Dirk
  Weissenborn}, \bibinfo{person}{Xiaohua Zhai}, \bibinfo{person}{Thomas
  Unterthiner}, \bibinfo{person}{Mostafa Dehghani}, \bibinfo{person}{Matthias
  Minderer}, \bibinfo{person}{Georg Heigold}, \bibinfo{person}{Sylvain Gelly},
  {et~al\mbox{.}}} \bibinfo{year}{2020}\natexlab{}.
\newblock \showarticletitle{An Image is Worth 16x16 Words: Transformers for
  Image Recognition at Scale}. In \bibinfo{booktitle}{\emph{International
  Conference on Learning Representations}}.
\newblock


\bibitem[Eigen et~al\mbox{.}(2014)]%
        {silog}
\bibfield{author}{\bibinfo{person}{David Eigen}, \bibinfo{person}{Christian
  Puhrsch}, {and} \bibinfo{person}{Rob Fergus}.}
  \bibinfo{year}{2014}\natexlab{}.
\newblock \showarticletitle{Depth map prediction from a single image using a
  multi-scale deep network}. In \bibinfo{booktitle}{\emph{Advances in neural
  information processing systems}}, Vol.~\bibinfo{volume}{27}.
  \bibinfo{pages}{2366--2374}.
\newblock


\bibitem[Fu et~al\mbox{.}(2018)]%
        {n15}
\bibfield{author}{\bibinfo{person}{Huan Fu}, \bibinfo{person}{Mingming Gong},
  \bibinfo{person}{Chaohui Wang}, \bibinfo{person}{Kayhan Batmanghelich}, {and}
  \bibinfo{person}{Dacheng Tao}.} \bibinfo{year}{2018}\natexlab{}.
\newblock \showarticletitle{Deep ordinal regression network for monocular depth
  estimation}. In \bibinfo{booktitle}{\emph{Proceedings of the IEEE/CVF
  Conference on Computer Vision and Pattern Recognition (CVPR)}}.
  \bibinfo{pages}{2002--2011}.
\newblock


\bibitem[Garg et~al\mbox{.}(2016)]%
        {photometricloss}
\bibfield{author}{\bibinfo{person}{Ravi Garg}, \bibinfo{person}{Vijay~Kumar
  Bg}, \bibinfo{person}{Gustavo Carneiro}, {and} \bibinfo{person}{Ian Reid}.}
  \bibinfo{year}{2016}\natexlab{}.
\newblock \showarticletitle{Unsupervised cnn for single view depth estimation:
  Geometry to the rescue}. In \bibinfo{booktitle}{\emph{European Conference on
  Computer Vision (ECCV)}}. Springer, \bibinfo{pages}{740--756}.
\newblock


\bibitem[Geiger et~al\mbox{.}(2013)]%
        {kitti}
\bibfield{author}{\bibinfo{person}{Andreas Geiger}, \bibinfo{person}{Philip
  Lenz}, \bibinfo{person}{Christoph Stiller}, {and} \bibinfo{person}{Raquel
  Urtasun}.} \bibinfo{year}{2013}\natexlab{}.
\newblock \showarticletitle{Vision meets robotics: The kitti dataset}.
\newblock \bibinfo{journal}{\emph{The International Journal of Robotics
  Research}} \bibinfo{volume}{32}, \bibinfo{number}{11} (\bibinfo{year}{2013}),
  \bibinfo{pages}{1231--1237}.
\newblock


\bibitem[Godard et~al\mbox{.}(2019)]%
        {monodepth2}
\bibfield{author}{\bibinfo{person}{C. Godard}, \bibinfo{person}{O. Aodha},
  \bibinfo{person}{M. Firman}, {and} \bibinfo{person}{G. Brostow}.}
  \bibinfo{year}{2019}\natexlab{}.
\newblock \showarticletitle{Digging into self-supervised monocular depth
  estimation}. In \bibinfo{booktitle}{\emph{Proceedings of the IEEE/CVF
  International Conference on Computer Vision (ICCV)}}.
  \bibinfo{pages}{3828--3838}.
\newblock


\bibitem[Guizilini et~al\mbox{.}(2021)]%
        {sans}
\bibfield{author}{\bibinfo{person}{Vitor Guizilini}, \bibinfo{person}{Rares
  Ambrus}, \bibinfo{person}{Wolfram Burgard}, {and} \bibinfo{person}{Adrien
  Gaidon}.} \bibinfo{year}{2021}\natexlab{}.
\newblock \showarticletitle{Sparse Auxiliary Networks for Unified Monocular
  Depth Prediction and Completion}. In \bibinfo{booktitle}{\emph{Proceedings of
  the IEEE/CVF Conference on Computer Vision and Pattern Recognition (CVPR)}}.
  \bibinfo{pages}{11078--11088}.
\newblock


\bibitem[Guizilini et~al\mbox{.}(2020a)]%
        {ddad}
\bibfield{author}{\bibinfo{person}{Vitor Guizilini}, \bibinfo{person}{Rares
  Ambrus}, \bibinfo{person}{Sudeep Pillai}, \bibinfo{person}{Allan Raventos},
  {and} \bibinfo{person}{Adrien Gaidon}.} \bibinfo{year}{2020}\natexlab{a}.
\newblock \showarticletitle{3D Packing for Self-Supervised Monocular Depth
  Estimation}. In \bibinfo{booktitle}{\emph{Proceedings of the IEEE/CVF
  Conference on Computer Vision and Pattern Recognition (CVPR)}}.
  \bibinfo{pages}{2482--2491}.
\newblock


\bibitem[Guizilini et~al\mbox{.}(2020b)]%
        {packnet-sfm}
\bibfield{author}{\bibinfo{person}{Vitor Guizilini}, \bibinfo{person}{Rares
  Ambrus}, \bibinfo{person}{Sudeep Pillai}, \bibinfo{person}{Allan Raventos},
  {and} \bibinfo{person}{Adrien Gaidon}.} \bibinfo{year}{2020}\natexlab{b}.
\newblock \showarticletitle{3D Packing for Self-Supervised Monocular Depth
  Estimation}. In \bibinfo{booktitle}{\emph{Proceedings of the IEEE/CVF
  Conference on Computer Vision and Pattern Recognition (CVPR)}}.
  \bibinfo{pages}{2482--2491}.
\newblock


\bibitem[Guizilini et~al\mbox{.}(2020c)]%
        {packnet}
\bibfield{author}{\bibinfo{person}{Vitor Guizilini}, \bibinfo{person}{Rares
  Ambrus}, \bibinfo{person}{Sudeep Pillai}, \bibinfo{person}{Allan Raventos},
  {and} \bibinfo{person}{Adrien Gaidon}.} \bibinfo{year}{2020}\natexlab{c}.
\newblock \showarticletitle{3d packing for self-supervised monocular depth
  estimation}. In \bibinfo{booktitle}{\emph{Proceedings of the IEEE/CVF
  Conference on Computer Vision and Pattern Recognition (CVPR)}}.
  \bibinfo{pages}{2485--2494}.
\newblock


\bibitem[Guizilini et~al\mbox{.}(2022)]%
        {fsm}
\bibfield{author}{\bibinfo{person}{Vitor Guizilini}, \bibinfo{person}{Igor
  Vasiljevic}, \bibinfo{person}{Rares Ambrus}, \bibinfo{person}{Greg
  Shakhnarovich}, {and} \bibinfo{person}{Adrien Gaidon}.}
  \bibinfo{year}{2022}\natexlab{}.
\newblock \showarticletitle{Full Surround Monodepth From Multiple Cameras}.
\newblock \bibinfo{journal}{\emph{IEEE Robotics and Automation Letters}}
  \bibinfo{volume}{7}, \bibinfo{number}{2} (\bibinfo{year}{2022}),
  \bibinfo{pages}{5397--5404}.
\newblock


\bibitem[Han et~al\mbox{.}(2022)]%
        {transdssl}
\bibfield{author}{\bibinfo{person}{Daechan Han}, \bibinfo{person}{Jeongmin
  Shin}, \bibinfo{person}{Namil Kim}, \bibinfo{person}{Soonmin Hwang}, {and}
  \bibinfo{person}{Yukyung Choi}.} \bibinfo{year}{2022}\natexlab{}.
\newblock \showarticletitle{Transdssl: Transformer based depth estimation via
  self-supervised learning}.
\newblock \bibinfo{journal}{\emph{IEEE Robotics and Automation Letters}}
  \bibinfo{volume}{7}, \bibinfo{number}{4} (\bibinfo{year}{2022}),
  \bibinfo{pages}{10969--10976}.
\newblock


\bibitem[Hartigan and Wong(1979)]%
        {kmeans}
\bibfield{author}{\bibinfo{person}{John~A Hartigan} {and}
  \bibinfo{person}{Manchek~A Wong}.} \bibinfo{year}{1979}\natexlab{}.
\newblock \showarticletitle{Algorithm AS 136: A k-means clustering algorithm}.
\newblock \bibinfo{journal}{\emph{Journal of the royal statistical society.
  series c (applied statistics)}} \bibinfo{volume}{28}, \bibinfo{number}{1}
  (\bibinfo{year}{1979}), \bibinfo{pages}{100--108}.
\newblock


\bibitem[He et~al\mbox{.}(2016)]%
        {resnet}
\bibfield{author}{\bibinfo{person}{Kaiming He}, \bibinfo{person}{Xiangyu
  Zhang}, \bibinfo{person}{Shaoqing Ren}, {and} \bibinfo{person}{Jian Sun}.}
  \bibinfo{year}{2016}\natexlab{}.
\newblock \showarticletitle{Deep residual learning for image recognition}. In
  \bibinfo{booktitle}{\emph{Proceedings of the IEEE/CVF Conference on Computer
  Vision and Pattern Recognition (CVPR)}}. \bibinfo{pages}{770--778}.
\newblock


\bibitem[Ho et~al\mbox{.}(2020)]%
        {ddpm}
\bibfield{author}{\bibinfo{person}{Jonathan Ho}, \bibinfo{person}{Ajay Jain},
  {and} \bibinfo{person}{Pieter Abbeel}.} \bibinfo{year}{2020}\natexlab{}.
\newblock \showarticletitle{Denoising diffusion probabilistic models}.
\newblock \bibinfo{journal}{\emph{Advances in Neural Information Processing
  Systems}}  \bibinfo{volume}{33} (\bibinfo{year}{2020}),
  \bibinfo{pages}{6840--6851}.
\newblock


\bibitem[Ho and Salimans(2022)]%
        {classfreeddpm}
\bibfield{author}{\bibinfo{person}{Jonathan Ho} {and} \bibinfo{person}{Tim
  Salimans}.} \bibinfo{year}{2022}\natexlab{}.
\newblock \showarticletitle{Classifier-free diffusion guidance}.
\newblock \bibinfo{journal}{\emph{arXiv preprint arXiv:2207.12598}}
  (\bibinfo{year}{2022}).
\newblock


\bibitem[Ho et~al\mbox{.}({[n.\,d.]})]%
        {difvideo}
\bibfield{author}{\bibinfo{person}{Jonathan Ho}, \bibinfo{person}{Tim
  Salimans}, \bibinfo{person}{Alexey~A Gritsenko}, \bibinfo{person}{William
  Chan}, \bibinfo{person}{Mohammad Norouzi}, {and} \bibinfo{person}{David~J
  Fleet}.} \bibinfo{year}{[n.\,d.]}\natexlab{}.
\newblock \showarticletitle{Video Diffusion Models}. In
  \bibinfo{booktitle}{\emph{Advances in Neural Information Processing
  Systems}}.
\newblock


\bibitem[Hu et~al\mbox{.}(2019)]%
        {wacv19loss}
\bibfield{author}{\bibinfo{person}{Junjie Hu}, \bibinfo{person}{Mete Ozay},
  \bibinfo{person}{Yan Zhang}, {and} \bibinfo{person}{Takayuki Okatani}.}
  \bibinfo{year}{2019}\natexlab{}.
\newblock \showarticletitle{Revisiting Single Image Depth Estimation: Toward
  Higher Resolution Maps With Accurate Object Boundaries}. In
  \bibinfo{booktitle}{\emph{2019 IEEE Winter Conference on Applications of
  Computer Vision (WACV)}}. \bibinfo{pages}{1043--1051}.
\newblock


\bibitem[Ignatov et~al\mbox{.}(2021)]%
        {mai21report}
\bibfield{author}{\bibinfo{person}{Andrey Ignatov}, \bibinfo{person}{Grigory
  Malivenko}, \bibinfo{person}{David Plowman}, \bibinfo{person}{Samarth
  Shukla}, {and} \bibinfo{person}{Radu Timofte}.}
  \bibinfo{year}{2021}\natexlab{}.
\newblock \showarticletitle{Fast and Accurate Single-Image Depth Estimation on
  Mobile Devices, Mobile AI 2021 Challenge: Report}. In
  \bibinfo{booktitle}{\emph{Proceedings of the IEEE/CVF Conference on Computer
  Vision and Pattern Recognition (CVPR) Workshops}}.
  \bibinfo{pages}{2545--2557}.
\newblock


\bibitem[Ignatov et~al\mbox{.}(2022)]%
        {mai22report}
\bibfield{author}{\bibinfo{person}{Andrey Ignatov}, \bibinfo{person}{Grigory
  Malivenko}, \bibinfo{person}{Radu Timofte}, \bibinfo{person}{Lukasz
  Treszczotko}, \bibinfo{person}{Xin Chang}, \bibinfo{person}{Piotr Ksiazek},
  \bibinfo{person}{Michal Lopuszynski}, \bibinfo{person}{Maciej Pioro},
  \bibinfo{person}{Rafal Rudnicki}, \bibinfo{person}{Maciej Smyl},
  {et~al\mbox{.}}} \bibinfo{year}{2022}\natexlab{}.
\newblock \showarticletitle{Efficient single-image depth estimation on mobile
  devices, mobile AI \& AIM 2022 challenge: report}. In
  \bibinfo{booktitle}{\emph{European Conference on Computer Vision}}. Springer,
  \bibinfo{pages}{71--91}.
\newblock


\bibitem[Lee et~al\mbox{.}(2019)]%
        {bts}
\bibfield{author}{\bibinfo{person}{Jin~Han Lee}, \bibinfo{person}{Myung-Kyu
  Han}, \bibinfo{person}{Dong~Wook Ko}, {and} \bibinfo{person}{Il~Hong Suh}.}
  \bibinfo{year}{2019}\natexlab{}.
\newblock \showarticletitle{From big to small: Multi-scale local planar
  guidance for monocular depth estimation}.
\newblock \bibinfo{journal}{\emph{arXiv preprint arXiv:1907.10326}}
  (\bibinfo{year}{2019}).
\newblock


\bibitem[Lin et~al\mbox{.}(2017b)]%
        {FFM1}
\bibfield{author}{\bibinfo{person}{Guosheng Lin}, \bibinfo{person}{Anton
  Milan}, \bibinfo{person}{Chunhua Shen}, {and} \bibinfo{person}{Ian Reid}.}
  \bibinfo{year}{2017}\natexlab{b}.
\newblock \showarticletitle{Refinenet: Multi-path refinement networks for
  high-resolution semantic segmentation}. In
  \bibinfo{booktitle}{\emph{Proceedings of the IEEE/CVF Conference on Computer
  Vision and Pattern Recognition (CVPR)}}. \bibinfo{pages}{1925--1934}.
\newblock


\bibitem[Lin et~al\mbox{.}(2017a)]%
        {FFM2}
\bibfield{author}{\bibinfo{person}{Tsung-Yi Lin}, \bibinfo{person}{Piotr
  Doll{\'a}r}, \bibinfo{person}{Ross Girshick}, \bibinfo{person}{Kaiming He},
  \bibinfo{person}{Bharath Hariharan}, {and} \bibinfo{person}{Serge Belongie}.}
  \bibinfo{year}{2017}\natexlab{a}.
\newblock \showarticletitle{Feature pyramid networks for object detection}. In
  \bibinfo{booktitle}{\emph{Proceedings of the IEEE/CVF Conference on Computer
  Vision and Pattern Recognition (CVPR)}}. \bibinfo{pages}{2117--2125}.
\newblock


\bibitem[Loshchilov and Hutter(2019)]%
        {weightdecay_adamw}
\bibfield{author}{\bibinfo{person}{Ilya Loshchilov} {and}
  \bibinfo{person}{Frank Hutter}.} \bibinfo{year}{2019}\natexlab{}.
\newblock \showarticletitle{Decoupled Weight Decay Regularization}.
\newblock  (\bibinfo{year}{2019}).
\newblock


\bibitem[Lyu et~al\mbox{.}(2021)]%
        {DifferentiableDepth}
\bibfield{author}{\bibinfo{person}{Zhaoyang Lyu}, \bibinfo{person}{Zhifeng
  Kong}, \bibinfo{person}{Xudong Xu}, \bibinfo{person}{Liang Pan}, {and}
  \bibinfo{person}{Dahua Lin}.} \bibinfo{year}{2021}\natexlab{}.
\newblock \showarticletitle{Differentiable Diffusion for Dense Depth Estimation
  From Multi-View Images}. In \bibinfo{booktitle}{\emph{Proceedings of the
  IEEE/CVF Conference on Computer Vision and Pattern Recognition (CVPR)}}.
  \bibinfo{pages}{8912--8921}.
\newblock


\bibitem[Nichol and Dhariwal(2021)]%
        {improvedddpm}
\bibfield{author}{\bibinfo{person}{Alexander~Quinn Nichol} {and}
  \bibinfo{person}{Prafulla Dhariwal}.} \bibinfo{year}{2021}\natexlab{}.
\newblock \showarticletitle{Improved denoising diffusion probabilistic models}.
  In \bibinfo{booktitle}{\emph{International Conference on Machine Learning}}.
  PMLR, \bibinfo{pages}{8162--8171}.
\newblock


\bibitem[Ranftl et~al\mbox{.}(2021)]%
        {dpt}
\bibfield{author}{\bibinfo{person}{Ren{\'e} Ranftl}, \bibinfo{person}{Alexey
  Bochkovskiy}, {and} \bibinfo{person}{Vladlen Koltun}.}
  \bibinfo{year}{2021}\natexlab{}.
\newblock \showarticletitle{Vision transformers for dense prediction}. In
  \bibinfo{booktitle}{\emph{Proceedings of the IEEE/CVF International
  Conference on Computer Vision (ICCV)}}. \bibinfo{pages}{12179--12188}.
\newblock


\bibitem[Ranftl et~al\mbox{.}(2020)]%
        {midas}
\bibfield{author}{\bibinfo{person}{Ren{\'e} Ranftl}, \bibinfo{person}{Katrin
  Lasinger}, \bibinfo{person}{David Hafner}, \bibinfo{person}{Konrad
  Schindler}, {and} \bibinfo{person}{Vladlen Koltun}.}
  \bibinfo{year}{2020}\natexlab{}.
\newblock \showarticletitle{Towards robust monocular depth estimation: Mixing
  datasets for zero-shot cross-dataset transfer}.
\newblock \bibinfo{journal}{\emph{IEEE transactions on pattern analysis and
  machine intelligence}} \bibinfo{volume}{44}, \bibinfo{number}{03}
  (\bibinfo{year}{2020}), \bibinfo{pages}{1623--1637}.
\newblock


\bibitem[Rombach et~al\mbox{.}(2022)]%
        {stable}
\bibfield{author}{\bibinfo{person}{Robin Rombach}, \bibinfo{person}{Andreas
  Blattmann}, \bibinfo{person}{Dominik Lorenz}, \bibinfo{person}{Patrick
  Esser}, {and} \bibinfo{person}{Bj\"orn Ommer}.}
  \bibinfo{year}{2022}\natexlab{}.
\newblock \showarticletitle{High-Resolution Image Synthesis With Latent
  Diffusion Models}. In \bibinfo{booktitle}{\emph{Proceedings of the IEEE/CVF
  Conference on Computer Vision and Pattern Recognition (CVPR)}}.
  \bibinfo{pages}{10684--10695}.
\newblock


\bibitem[Saharia et~al\mbox{.}(2022)]%
        {palette}
\bibfield{author}{\bibinfo{person}{Chitwan Saharia}, \bibinfo{person}{William
  Chan}, \bibinfo{person}{Huiwen Chang}, \bibinfo{person}{Chris Lee},
  \bibinfo{person}{Jonathan Ho}, \bibinfo{person}{Tim Salimans},
  \bibinfo{person}{David Fleet}, {and} \bibinfo{person}{Mohammad Norouzi}.}
  \bibinfo{year}{2022}\natexlab{}.
\newblock \showarticletitle{Palette: Image-to-image diffusion models}. In
  \bibinfo{booktitle}{\emph{ACM SIGGRAPH 2022 Conference Proceedings}}.
  \bibinfo{pages}{1--10}.
\newblock


\bibitem[Salimans and Ho(2022)]%
        {fastdif_iclr22}
\bibfield{author}{\bibinfo{person}{Tim Salimans} {and}
  \bibinfo{person}{Jonathan Ho}.} \bibinfo{year}{2022}\natexlab{}.
\newblock \showarticletitle{Progressive Distillation for Fast Sampling of
  Diffusion Models}. In \bibinfo{booktitle}{\emph{International Conference on
  Learning Representations}}.
\newblock


\bibitem[Sch\"{o}nberger and Frahm(2016)]%
        {colmapsfm}
\bibfield{author}{\bibinfo{person}{Johannes~Lutz Sch\"{o}nberger} {and}
  \bibinfo{person}{Jan-Michael Frahm}.} \bibinfo{year}{2016}\natexlab{}.
\newblock \showarticletitle{Structure-from-Motion Revisited}. In
  \bibinfo{booktitle}{\emph{Proceedings of the IEEE/CVF Conference on Computer
  Vision and Pattern Recognition (CVPR)}}. \bibinfo{pages}{4104--4113}.
\newblock


\bibitem[Shu et~al\mbox{.}(2020)]%
        {shu}
\bibfield{author}{\bibinfo{person}{Chang Shu}, \bibinfo{person}{Kun Yu},
  \bibinfo{person}{Zhixiang Duan}, {and} \bibinfo{person}{Kuiyuan Yang}.}
  \bibinfo{year}{2020}\natexlab{}.
\newblock \showarticletitle{Feature-Metric Loss for Self-Supervised Learning of
  Depth and Egomotion}. In \bibinfo{booktitle}{\emph{European Conference on
  Computer Vision (ECCV)}}. Springer, \bibinfo{pages}{572–588}.
\newblock


\bibitem[Sohl-Dickstein et~al\mbox{.}(2015)]%
        {ddpmini}
\bibfield{author}{\bibinfo{person}{Jascha Sohl-Dickstein},
  \bibinfo{person}{Eric Weiss}, \bibinfo{person}{Niru Maheswaranathan}, {and}
  \bibinfo{person}{Surya Ganguli}.} \bibinfo{year}{2015}\natexlab{}.
\newblock \showarticletitle{Deep unsupervised learning using nonequilibrium
  thermodynamics}. In \bibinfo{booktitle}{\emph{International Conference on
  Machine Learning}}. PMLR, \bibinfo{pages}{2256--2265}.
\newblock


\bibitem[Song et~al\mbox{.}(2020a)]%
        {ddim}
\bibfield{author}{\bibinfo{person}{Jiaming Song}, \bibinfo{person}{Chenlin
  Meng}, {and} \bibinfo{person}{Stefano Ermon}.}
  \bibinfo{year}{2020}\natexlab{a}.
\newblock \showarticletitle{Denoising Diffusion Implicit Models}. In
  \bibinfo{booktitle}{\emph{International Conference on Learning
  Representations}}.
\newblock


\bibitem[Song et~al\mbox{.}(2020b)]%
        {scoredif}
\bibfield{author}{\bibinfo{person}{Yang Song}, \bibinfo{person}{Jascha
  Sohl-Dickstein}, \bibinfo{person}{Diederik~P Kingma},
  \bibinfo{person}{Abhishek Kumar}, \bibinfo{person}{Stefano Ermon}, {and}
  \bibinfo{person}{Ben Poole}.} \bibinfo{year}{2020}\natexlab{b}.
\newblock \showarticletitle{Score-based generative modeling through stochastic
  differential equations}.
\newblock  (\bibinfo{year}{2020}).
\newblock


\bibitem[Vaswani et~al\mbox{.}(2017)]%
        {transformer}
\bibfield{author}{\bibinfo{person}{Ashish Vaswani}, \bibinfo{person}{Noam
  Shazeer}, \bibinfo{person}{Niki Parmar}, \bibinfo{person}{Jakob Uszkoreit},
  \bibinfo{person}{Llion Jones}, \bibinfo{person}{Aidan~N Gomez},
  \bibinfo{person}{{\L}ukasz Kaiser}, {and} \bibinfo{person}{Illia
  Polosukhin}.} \bibinfo{year}{2017}\natexlab{}.
\newblock \showarticletitle{Attention is all you need}. In
  \bibinfo{booktitle}{\emph{Advances in neural information processing
  systems}}, Vol.~\bibinfo{volume}{30}.
\newblock


\bibitem[Wang et~al\mbox{.}(2022b)]%
        {planedepth}
\bibfield{author}{\bibinfo{person}{Ruoyu Wang}, \bibinfo{person}{Zehao Yu},
  {and} \bibinfo{person}{Shenghua Gao}.} \bibinfo{year}{2022}\natexlab{b}.
\newblock \showarticletitle{PlaneDepth: Plane-Based Self-Supervised Monocular
  Depth Estimation}.
\newblock \bibinfo{journal}{\emph{arXiv preprint arXiv:2210.01612}}
  (\bibinfo{year}{2022}).
\newblock


\bibitem[Wang et~al\mbox{.}(2021)]%
        {mai21}
\bibfield{author}{\bibinfo{person}{Yiran Wang}, \bibinfo{person}{Xingyi Li},
  \bibinfo{person}{Min Shi}, \bibinfo{person}{Ke Xian}, {and}
  \bibinfo{person}{Zhiguo Cao}.} \bibinfo{year}{2021}\natexlab{}.
\newblock \showarticletitle{Knowledge Distillation for Fast and Accurate
  Monocular Depth Estimation on Mobile Devices}. In
  \bibinfo{booktitle}{\emph{Proceedings of the IEEE/CVF Conference on Computer
  Vision and Pattern Recognition (CVPR) Workshops}}.
  \bibinfo{pages}{2457--2465}.
\newblock


\bibitem[Wang et~al\mbox{.}(2022a)]%
        {fmnet}
\bibfield{author}{\bibinfo{person}{Yiran Wang}, \bibinfo{person}{Zhiyu Pan},
  \bibinfo{person}{Xingyi Li}, \bibinfo{person}{Zhiguo Cao},
  \bibinfo{person}{Ke Xian}, {and} \bibinfo{person}{Jianming Zhang}.}
  \bibinfo{year}{2022}\natexlab{a}.
\newblock \showarticletitle{Less is More: Consistent Video Depth Estimation
  with Masked Frames Modeling}. In \bibinfo{booktitle}{\emph{Proceedings of the
  30th ACM International Conference on Multimedia}} (Lisboa, Portugal)
  \emph{(\bibinfo{series}{MM '22})}. \bibinfo{publisher}{Association for
  Computing Machinery}, \bibinfo{address}{New York, NY, USA},
  \bibinfo{pages}{6347–6358}.
\newblock
\showISBNx{9781450392037}
\urldef\tempurl%
\url{https://doi.org/10.1145/3503161.3547978}
\showDOI{\tempurl}


\bibitem[Wang et~al\mbox{.}(2023)]%
        {nvds}
\bibfield{author}{\bibinfo{person}{Yiran Wang}, \bibinfo{person}{Min Shi},
  \bibinfo{person}{Jiaqi Li}, \bibinfo{person}{Zihao Huang},
  \bibinfo{person}{Zhiguo Cao}, \bibinfo{person}{Jianming Zhang},
  \bibinfo{person}{Ke Xian}, {and} \bibinfo{person}{Guosheng Lin}.}
  \bibinfo{year}{2023}\natexlab{}.
\newblock \showarticletitle{Neural Video Depth Stabilizer}.
\newblock \bibinfo{journal}{\emph{arXiv preprint arXiv:2307.08695}}
  (\bibinfo{year}{2023}).
\newblock


\bibitem[Watson et~al\mbox{.}(2021)]%
        {manydepth}
\bibfield{author}{\bibinfo{person}{Jamie Watson}, \bibinfo{person}{Oisin
  Mac~Aodha}, \bibinfo{person}{Victor Prisacariu}, \bibinfo{person}{Gabriel
  Brostow}, {and} \bibinfo{person}{Michael Firman}.}
  \bibinfo{year}{2021}\natexlab{}.
\newblock \showarticletitle{The Temporal Opportunist: Self-Supervised
  Multi-Frame Monocular Depth}. In \bibinfo{booktitle}{\emph{Proceedings of the
  IEEE/CVF Conference on Computer Vision and Pattern Recognition (CVPR)}}.
  \bibinfo{pages}{1164--1174}.
\newblock


\bibitem[Wei et~al\mbox{.}(2023)]%
        {surround}
\bibfield{author}{\bibinfo{person}{Yi Wei}, \bibinfo{person}{Linqing Zhao},
  \bibinfo{person}{Wenzhao Zheng}, \bibinfo{person}{Zheng Zhu},
  \bibinfo{person}{Yongming Rao}, \bibinfo{person}{Guan Huang},
  \bibinfo{person}{Jiwen Lu}, {and} \bibinfo{person}{Jie Zhou}.}
  \bibinfo{year}{2023}\natexlab{}.
\newblock \showarticletitle{SurroundDepth: Entangling Surrounding Views for
  Self-Supervised Multi-Camera Depth Estimation}. In
  \bibinfo{booktitle}{\emph{Proceedings of The 6th Conference on Robot
  Learning}}, Vol.~\bibinfo{volume}{205}. \bibinfo{pages}{539--549}.
\newblock


\bibitem[Wolleb et~al\mbox{.}(2022)]%
        {difseg_medical}
\bibfield{author}{\bibinfo{person}{Julia Wolleb}, \bibinfo{person}{Robin
  Sandk{\"u}hler}, \bibinfo{person}{Florentin Bieder},
  \bibinfo{person}{Philippe Valmaggia}, {and} \bibinfo{person}{Philippe~C
  Cattin}.} \bibinfo{year}{2022}\natexlab{}.
\newblock \showarticletitle{Diffusion models for implicit image segmentation
  ensembles}. In \bibinfo{booktitle}{\emph{International Conference on Medical
  Imaging with Deep Learning}}. PMLR, \bibinfo{pages}{1336--1348}.
\newblock


\bibitem[Xian et~al\mbox{.}(2020)]%
        {kexian2020}
\bibfield{author}{\bibinfo{person}{Ke Xian}, \bibinfo{person}{Jianming Zhang},
  \bibinfo{person}{Oliver Wang}, \bibinfo{person}{Long Mai},
  \bibinfo{person}{Zhe Lin}, {and} \bibinfo{person}{Zhiguo Cao}.}
  \bibinfo{year}{2020}\natexlab{}.
\newblock \showarticletitle{Structure-Guided Ranking Loss for Single Image
  Depth Prediction}. In \bibinfo{booktitle}{\emph{Proceedings of the IEEE/CVF
  Conference on Computer Vision and Pattern Recognition (CVPR)}}.
  \bibinfo{pages}{608--617}.
\newblock


\bibitem[Xie et~al\mbox{.}(2017)]%
        {resnext}
\bibfield{author}{\bibinfo{person}{Saining Xie}, \bibinfo{person}{Ross
  Girshick}, \bibinfo{person}{Piotr Doll{\'a}r}, \bibinfo{person}{Zhuowen Tu},
  {and} \bibinfo{person}{Kaiming He}.} \bibinfo{year}{2017}\natexlab{}.
\newblock \showarticletitle{Aggregated residual transformations for deep neural
  networks}. In \bibinfo{booktitle}{\emph{Proceedings of the IEEE/CVF
  Conference on Computer Vision and Pattern Recognition (CVPR)}}.
  \bibinfo{pages}{1492--1500}.
\newblock


\bibitem[Xu et~al\mbox{.}(2022)]%
        {mcdp}
\bibfield{author}{\bibinfo{person}{Jialei Xu}, \bibinfo{person}{Xianming Liu},
  \bibinfo{person}{Yuanchao Bai}, \bibinfo{person}{Junjun Jiang},
  \bibinfo{person}{Kaixuan Wang}, \bibinfo{person}{Xiaozhi Chen}, {and}
  \bibinfo{person}{Xiangyang Ji}.} \bibinfo{year}{2022}\natexlab{}.
\newblock \showarticletitle{Multi-Camera Collaborative Depth Prediction via
  Consistent Structure Estimation}. In \bibinfo{booktitle}{\emph{Proceedings of
  the 30th ACM International Conference on Multimedia}} (Lisboa, Portugal)
  \emph{(\bibinfo{series}{MM '22})}. \bibinfo{publisher}{Association for
  Computing Machinery}, \bibinfo{address}{New York, NY, USA},
  \bibinfo{pages}{2730–2738}.
\newblock
\showISBNx{9781450392037}
\urldef\tempurl%
\url{https://doi.org/10.1145/3503161.3548394}
\showDOI{\tempurl}


\bibitem[Yin et~al\mbox{.}(2019)]%
        {vnl}
\bibfield{author}{\bibinfo{person}{Wei Yin}, \bibinfo{person}{Yifan Liu},
  \bibinfo{person}{Chunhua Shen}, {and} \bibinfo{person}{Youliang Yan}.}
  \bibinfo{year}{2019}\natexlab{}.
\newblock \showarticletitle{Enforcing geometric constraints of virtual normal
  for depth prediction}. In \bibinfo{booktitle}{\emph{Proceedings of the
  IEEE/CVF International Conference on Computer Vision (ICCV)}}.
  \bibinfo{pages}{5684--5693}.
\newblock


\bibitem[Yuan et~al\mbox{.}(2022)]%
        {newcrfs}
\bibfield{author}{\bibinfo{person}{Weihao Yuan}, \bibinfo{person}{Xiaodong Gu},
  \bibinfo{person}{Zuozhuo Dai}, \bibinfo{person}{Siyu Zhu}, {and}
  \bibinfo{person}{Ping Tan}.} \bibinfo{year}{2022}\natexlab{}.
\newblock \showarticletitle{NeWCRFs: Neural Window Fully-connected CRFs for
  Monocular Depth Estimation}. In \bibinfo{booktitle}{\emph{Proceedings of the
  IEEE/CVF Conference on Computer Vision and Pattern Recognition (CVPR)}}.
  \bibinfo{pages}{3916--3925}.
\newblock


\end{thebibliography}
\clearpage
\appendix

\section{More details on DADP framework}
\label{sec:ss1}
\subsection{The Noise Predictor}

\label{sec:zs}
We adopt Unet as the \zs{} with multi-resolution attention~\cite{ddpmbeat} and BigGAN residual blocks~\cite{biggan}. Here we specify the details.

\begin{figure}[h]
    \centering
    \includegraphics[scale=0.6,trim=10 0 10 0,clip]{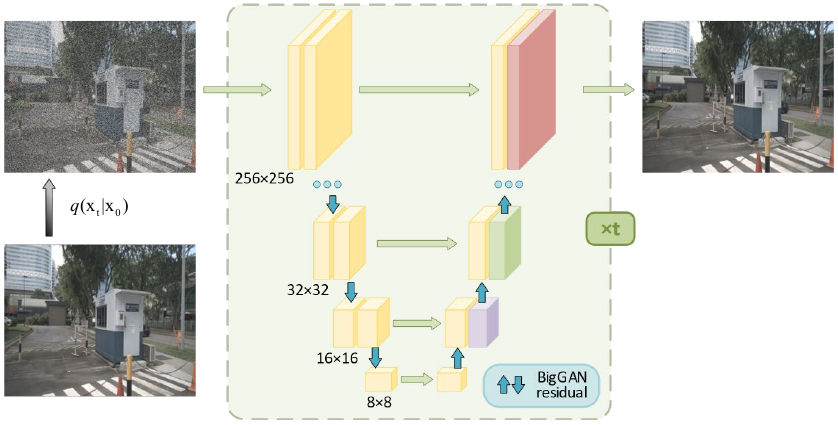}
    \caption{Architecture of the \zs{}. Some resolutions are omitted for simplicity. Blue arrows indicate the residual blocks in BigGAN~\cite{biggan} for up-/down-sampling. Two Resblocks~\cite{resnet} are used for each resolution.}
    \label{fig:unet}
    \vspace{-15pt}
\end{figure}

As shown in ~\reffig~\ref{fig:unet}, the \zs{} downsamples the input noisy image to $1/32$ of the original resolution. For each resolution, two Resblocks~\cite{resnet} are used for feature extraction. Linear layers are used in these blocks to embed the diffusion step $t$. We utilize multi-head self-attention~\cite{transformer} on resolutions of $32\times32$, $16\times16$, $8\times8$ to improve the representation ability. Besides, the simple up-sampling and down-sampling between different resolutions are replaced by the residual blocks in BigGAN~\cite{biggan}, which can reduce information loss and improve the generation quality.

\subsection{Object-guided Integrality Loss}
\label{sec:lossvis}
\begin{figure}[H]
    \centering
    \includegraphics[scale=0.38,trim=0 0 0 0,clip]{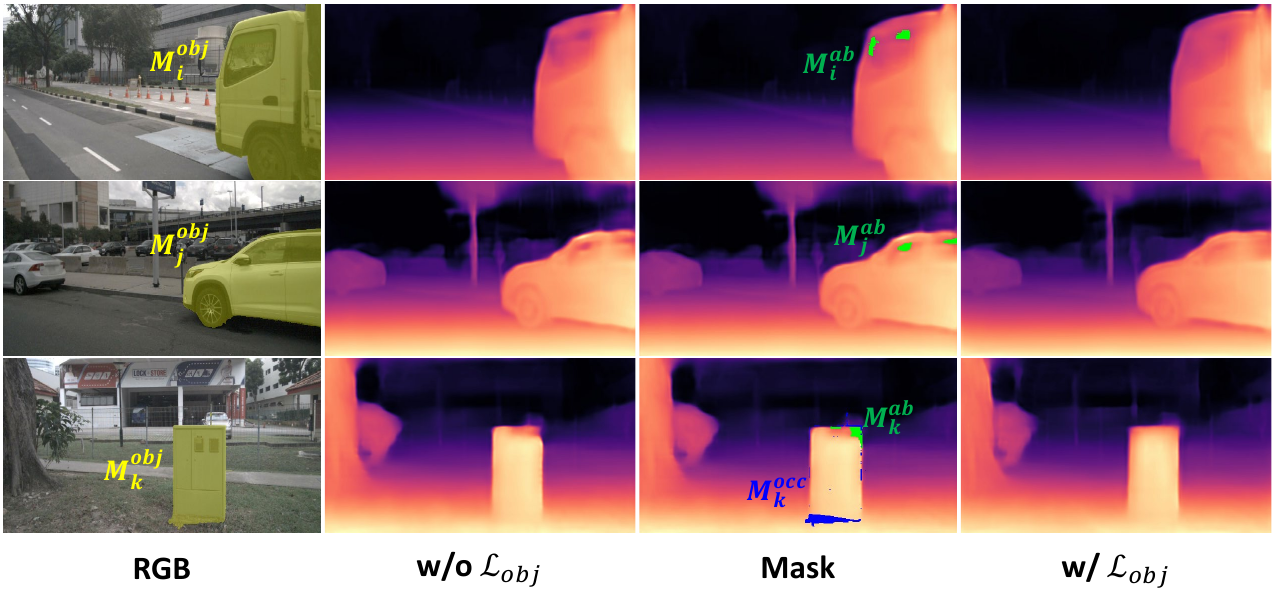}
    \caption{Masks in \loss{}. The yellow areas in the rgb images represent the object mask $\mathbf{M}^{obj}$ obtained by segmentation model~\cite{mask2former}. The green areas $\mathbf{M}^{ab}$ in the third column indicate the regions with abnormal depth variations. The blue area $\mathbf{M}^{occ}$ represents the excluded pixels with segmentation errors by k-means clustering~\cite{kmeans}. The indices $\textbf{i, j, k}$ denote specific objects in the three images. }
    \label{fig:lossvis}
\end{figure}

Here we explain different masks in the \loss{}. We visualize $\mathbf{M}^{obj},\mathbf{M}^{ab}$, and $\mathbf{M}^{occ}$ for interpretation in \reffig~\ref{fig:lossvis}. The indices $i,j$, and $k$ represent different objects in the input image. $\mathbf{M}^{ab}$ can effectively mark the abnormal depth areas in the certain object $\mathbf{M}^{obj}$. For the regions with complex occlusions or segmentation errors, such as the region beneath the $\mathbf{M}_k^{obj}$ in the third row, we also exclude them as the $\mathbf{M}_k^{occ}$ by k-means clustering~\cite{kmeans}.

\subsection{Affinity Invariant Loss}
\label{sec:afloss}
 For the training of the depth predictor~\cite{kexian2020,dpt,midas}, we also use the affinity invariant loss~\cite{midas,dpt} in addition to the proposed \loss{}. Here we briefly present the details. If we denote the original predicted depth and ground truth as $\mathbf{d}$ and $\mathbf{d}^{*}$, the
scale $s(\mathbf{d})$ and shift $t(\mathbf{d})$ of prediction $\mathbf{d}$ can be obtained as:
\begin{equation}
    t(\mathbf{d}) = median(\mathbf{d}), 
    s(\mathbf{d})=\frac{1}{M}\sum_{i=1}^M|\mathbf{d}_i-t(\mathbf{d}_i)|\,,
\end{equation}
where $M$ denotes the number of pixels. $t(\mathbf{d}^*)$ and $s(\mathbf{d}^*)$ can also be calculated similarly. 
Then the prediction $\mathbf{d}$ and the ground truth $\mathbf{d}^{*}$ are aligned to zero translation and unit scale:
\begin{equation}
    \hat{\mathbf{d}} = \frac{\mathbf{d}-t(\mathbf{d})}{s(\mathbf{d})},\hat{\mathbf{d}}^* = \frac{\mathbf{d}^*-t(\mathbf{d}^*)}{s(\mathbf{d}^*)}\;.
\end{equation}
$\hat{\mathbf{d}}$ and $\hat{\mathbf{d}}^{*}$ represent the aligned prediction and ground truth respectively. The affinity invariant loss $\mathcal{L}_{af}$ can be formulated as:
\begin{equation}
    \mathcal{L}_{af}\left(\hat{\mathbf{d}}, \hat{\mathbf{d}}^*\right)=\frac{1}{2 M} \sum_{i=1}^M \left|\hat{\mathbf{d}}_i-\hat{\mathbf{d}}_i^*\right|,
    \label{equ:loss2}
\end{equation}

\section{Depth Estimation Metrics}
\label{sec:metric}
We adopt the commonly-applied depth metrics defined as follows:
\begin{itemize}[leftmargin=*]
\item \textbf{Absolute relative error (Abs Rel):} $\frac{1}{|M|} \sum_{d \in M}\left|d-d^*\right| / d^* ;$
\item \textbf{Square relative error (Sq Rel):} $\frac{1}{|M|} \sum_{d \in M}\left\|d-d^*\right\|^2 / d^*$
\item \textbf{Root mean square error (RMSE):} $\sqrt{\frac{1}{|M|} \sum_{d \in M}\left\|d-d^*\right\|^2} ;$
\item \textbf{Accuracy with threshold t:} Percentage of $d_i$ such that $\\max(\frac{d_i}{d_i^*},\frac{d_i^*}{d_i}) = \delta<t\in\left[1.25, 1.25^2, 1.25^3\right]\,,$ 
\end{itemize}
where $M$ denotes numbers of pixels with valid depth annotation, $d_i$ and $d_i^*$ are estimated and ground truth depth of pixel $i$ respectively.

\section{More results for motivation review.}
\label{sec:supp_motivation}
\noindent \textbf{Data Augmentation and Weight Decay.} In Sec.~4.2 of main paper, we illustrate that data augmentations and weight decay (\textit{e.g.}, regularization) cannot solve the incomplete objects, concave areas, and artifacts for overfitting to sparse valid pixels. The related visual comparisons are given in the Fig.~3 of the main paper. Here we provide the corresponding quantitative metrics in ~\reftab{}~\ref{tab:aug}.

We find that data augmentations reduce the depth accuracy for the reason that those augmentations change the data distribution of \nus{} dataset~\cite{nus}. The random cropping sometimes leads to even sparser depth supervision. Our method improves the depth accuracy and enhances the spatial structural integrity of the predicted depth maps by fusing structure-aware features from the \zs{}.

\noindent \textbf{Features of CNN Encoders.} In Sec.~4.2 of the main paper, the noise predictor is replaced by widely-used CNN encoders~\cite{resnet,resnext}. We fuse the extracted features into the depth predictor. The visualization of those features in Fig.~4 of the main paper demonstrates the better representations of spatial structures for our \zs{}. Here we report visual and quantitative results of this experiment. 

\begin{figure}[h]
    \centering
    \includegraphics[scale=0.35,trim=0 0 0 0,clip]{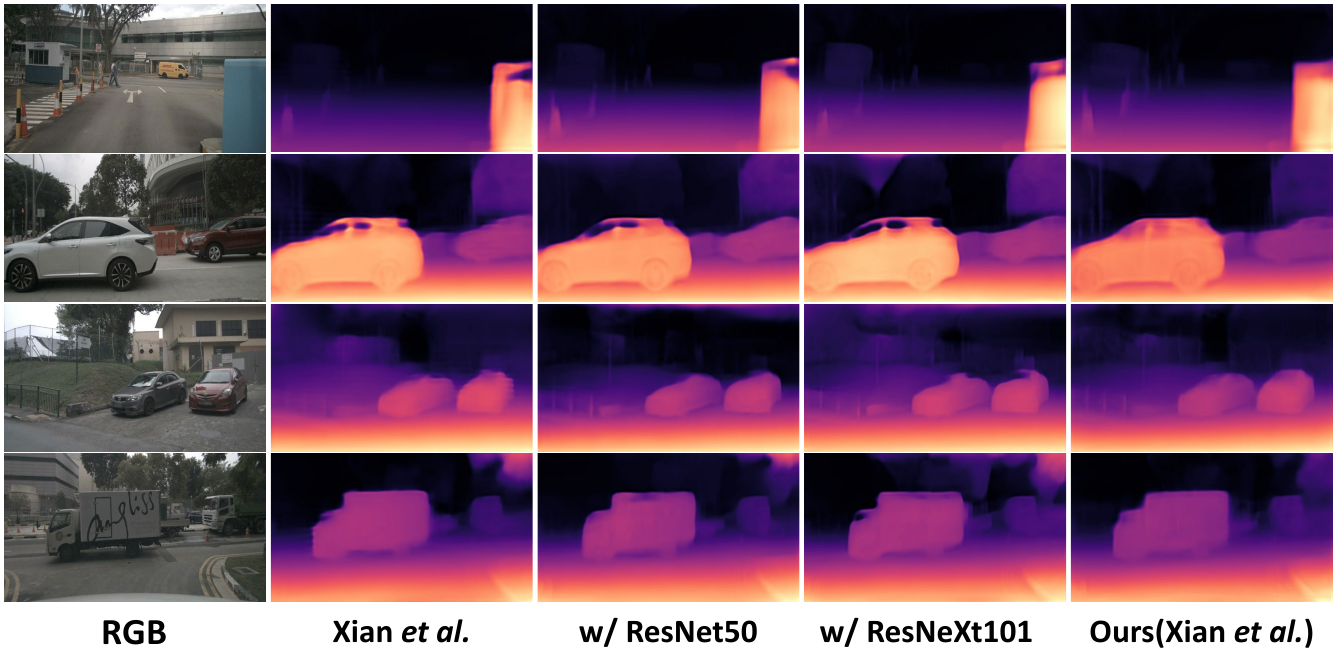}
    \caption{Visual comparisons with CNN encoders~\cite{resnet,resnext}. Our DADP effectively augments spatial structural integrality of depth predictions by \zs{} and diffusion model.}
    \label{fig:res}
\end{figure}

As shown in ~\reffig{}~\ref{fig:res} and ~\reftab{}~\ref{tab:cnnencoder}, along with the Fig.~4 in the main paper, we showcase the effectiveness of introducing structure-aware features from the \zs{} and diffusion model. The results demonstrate that the structure-aware features can significantly improve the spatial structural integrality of output depth.

\begin{table}[t]
  \setlength{\tabcolsep}{2pt}
  \caption{Comparisons of data augmentations (Aug) and weight decay (Reg) on \nus{} dataset~\cite{nus}.}
  \label{tab:aug}
  \resizebox{\columnwidth}{!}{
  \begin{tabular}{lcccccc} 
    \toprule
    Method & Abs Rel$\downarrow$ & Sq Rel$\downarrow$ & RMSE$\downarrow$ & $\delta_1$$\uparrow$ & $\delta_2$$\uparrow$ & $\delta_3$$\uparrow$ \\
    \midrule
    Midas~\cite{midas}     & $0.122$ & $1.106$ & $5.485$ & $0.844$ & $0.933$ & $0.964$\\
    $w/$ Reg               & $0.126$ & $1.154$ & $5.709$ & $0.840$ & $0.930$ & $0.963$\\
    $w/$ Aug               & $0.141$ & $1.334$ & $5.696$ & $0.816$ & $0.917$ & $0.957$\\
    $w/$ Aug\&Reg          & $0.141$ & $1.339$ & $5.705$ & $0.815$ & $0.919$ & $0.958$\\
    Ours(MiDaS)            & $\textbf{0.117}$ & $\textbf{1.084}$ & $\textbf{5.370}$ & $\textbf{0.856}$ & $\textbf{0.938}$ & $\textbf{0.967}$\\
    \bottomrule
\end{tabular}
}
\end{table}
\begin{table}[t]
  \setlength{\tabcolsep}{2pt}
  \caption{Comparisons with features of different CNN encoders on \nus{}~\cite{nus}. Best performance is in boldface. }
  \label{tab:cnnencoder}
  \resizebox{\columnwidth}{!}{
  \begin{tabular}{lcccccc} 
    \toprule
    Method & Abs Rel$\downarrow$ & Sq Rel$\downarrow$ & RMSE$\downarrow$ & $\delta_1$$\uparrow$ & $\delta_2$$\uparrow$ & $\delta_3$$\uparrow$ \\
    \midrule
    Xian \textit{et al.}~\cite{kexian2020}     & $0.183$ & $1.375$ & $6.266$ & $0.799$ & $0.913$ & $0.957$\\
    $w/$ ResNet50       & $0.142$ & $1.269$ & $6.054$ & $0.808$ & $0.917$ & $0.960$\\
    $w/$ ResNeXt101     & $0.140$ & $1.277$ & $\textbf{5.918}$ & $0.813$ & $0.920$ & $0.960$\\
    $w/$ \zs{}          & $\textbf{0.132}$ & $\textbf{1.141}$ & $5.951$ & $\textbf{0.824}$ & $\textbf{0.925}$ & $\textbf{0.963}$\\
    \bottomrule
\end{tabular}
}
\vspace{-10pt}
\end{table}

\section{Experiments on ~\ki{} dataset}
In Sec.~4.1 of the main paper, we mentioned that we also train our \sxf{} on ~\ki{} dataset~\cite{kitti} for sufficient comparisons and evaluations, following the train/test split as Eigen \textit{et al.}~\cite{silog,bts}.

\begin{table}[t]
  \setlength{\tabcolsep}{2pt}
  \caption{Comparisons with \sota{} methods on \ki{} dataset~\cite{kitti}. The first five rows contain results of self-supervised methods, while the last seven rows contain supervised approaches. Best performance is in boldface.}
  \label{tab:kitti}
  \resizebox{\columnwidth}{!}{
  \begin{tabular}{lcccccc} 
    \toprule
    Method & Abs Rel$\downarrow$ & Sq Rel$\downarrow$ & RMSE$\downarrow$ & $\delta_1$$\uparrow$ & $\delta_2$$\uparrow$ & $\delta_3$$\uparrow$ \\
    \midrule
    Monodepth2~\cite{monodepth2}          & $0.115$ & $0.882$ & $4.701$ & $0.879$ & $0.961$ & $0.982$\\
    Packnet-SFM~\cite{packnet-sfm}         & $0.107$ & $0.802$ & $4.538$ & $0.889$ & $0.962$ & $0.981$\\
    Shu \textit{et al.}~\cite{shu} & $0.088$ & $0.712$ & $4.137$ & $0.915$ & $0.965$ & $0.982$\\
    PlaneDepth~\cite{planedepth}         & $0.083$ & $0.533$ & $3.919$ & $0.913$ & $0.969$ & $0.985$\\
    ManyDepth~\cite{manydepth}           & $0.087$ & $0.685$ & $4.142$ & $0.920$ & $0.968$ & $0.983$\\
    \midrule
    DORN~\cite{n15}        & $0.072$ & $-$ & $2.626$ & $0.932$ & $0.984$ & $0.994$\\
    VNL~\cite{vnl}         & $0.072$ & $-$ & $3.258$ & $0.938$ & $0.990$ & $0.998$\\
    MiDaS~\cite{midas}  & $0.069$ & $0.280$ & $3.006$ & $0.949$ & $0.991$ & $0.998$\\
    Ours(MiDaS)         & $0.065$ & $0.253$ & $2.832$ & $0.955$ & $0.992$ & $0.998$\\
    BTS~\cite{bts}      & $0.059$ & $0.241$ & $2.756$ & $0.956$ & $0.993$ & $0.998$\\
    DPT~\cite{dpt}      & $0.062$ & $\textbf{0.222}$ & $\textbf{2.573}$ & $0.959$ & $0.995$ & $0.999$\\
    Ours(DPT)         & $\textbf{0.059}$ & $0.230$ & $2.661$ & $\textbf{0.965}$ & $\textbf{0.995}$ & $\textbf{0.999}$\\
    \bottomrule
\end{tabular}
}
\end{table}

Quantitative comparisons in ~\reftab{}~\ref{tab:kitti} show that our \sxf{} improves the depth accuracy than the depth predictors Midas~\cite{midas} and DPT~\cite{dpt} on \ki{} dataset~\cite{kitti}.

\section{More qualitative depth results}
Due to the sparsity of the depth annotations on autonomous driving scenarios~\cite{nus,ddad,kitti}, the quantitative metrics cannot fully reflect our improvements. Here we show more visual comparisons to demonstrate the effectiveness of our approach, especially for better spatial structural completeness of the predicted depth maps. 

\begin{figure*}[h]
    \centering
    \includegraphics[scale=0.525,trim=0 0 0 0,clip]{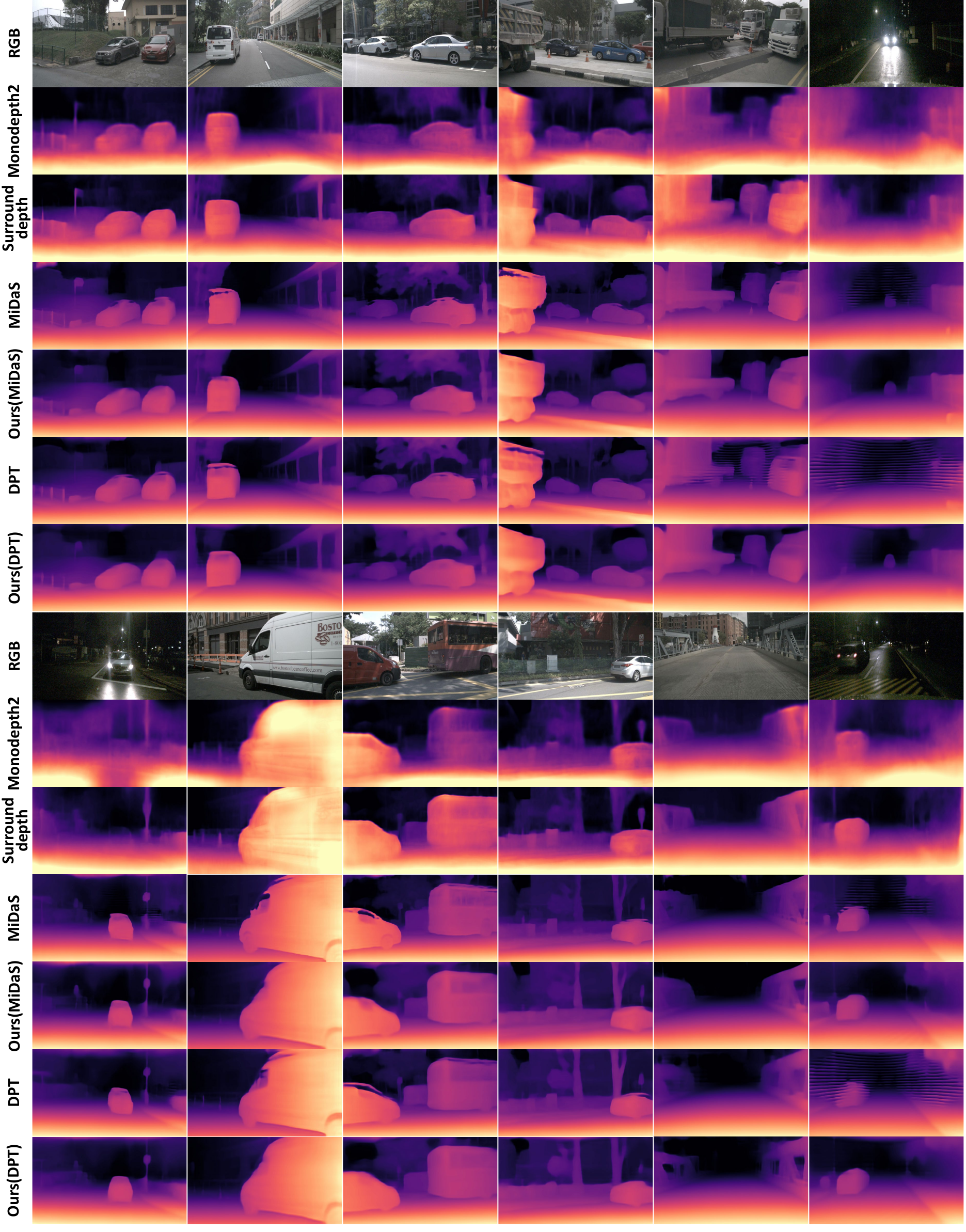}
    \vspace{-8pt}
    \caption{Visual comparisons on \nus{} dataset~\cite{nus}. we compare our \sxf{} framework with different depth predictors~\cite{midas,dpt} and \sota{} approaches~\cite{monodepth2,surround}. The results demonstrate the effectiveness of our plug-and-play manner.}
    \label{fig:suppnus}
\end{figure*}

\begin{figure*}[h]
    \centering
    \includegraphics[scale=0.60,trim=0 0 0 0,clip]{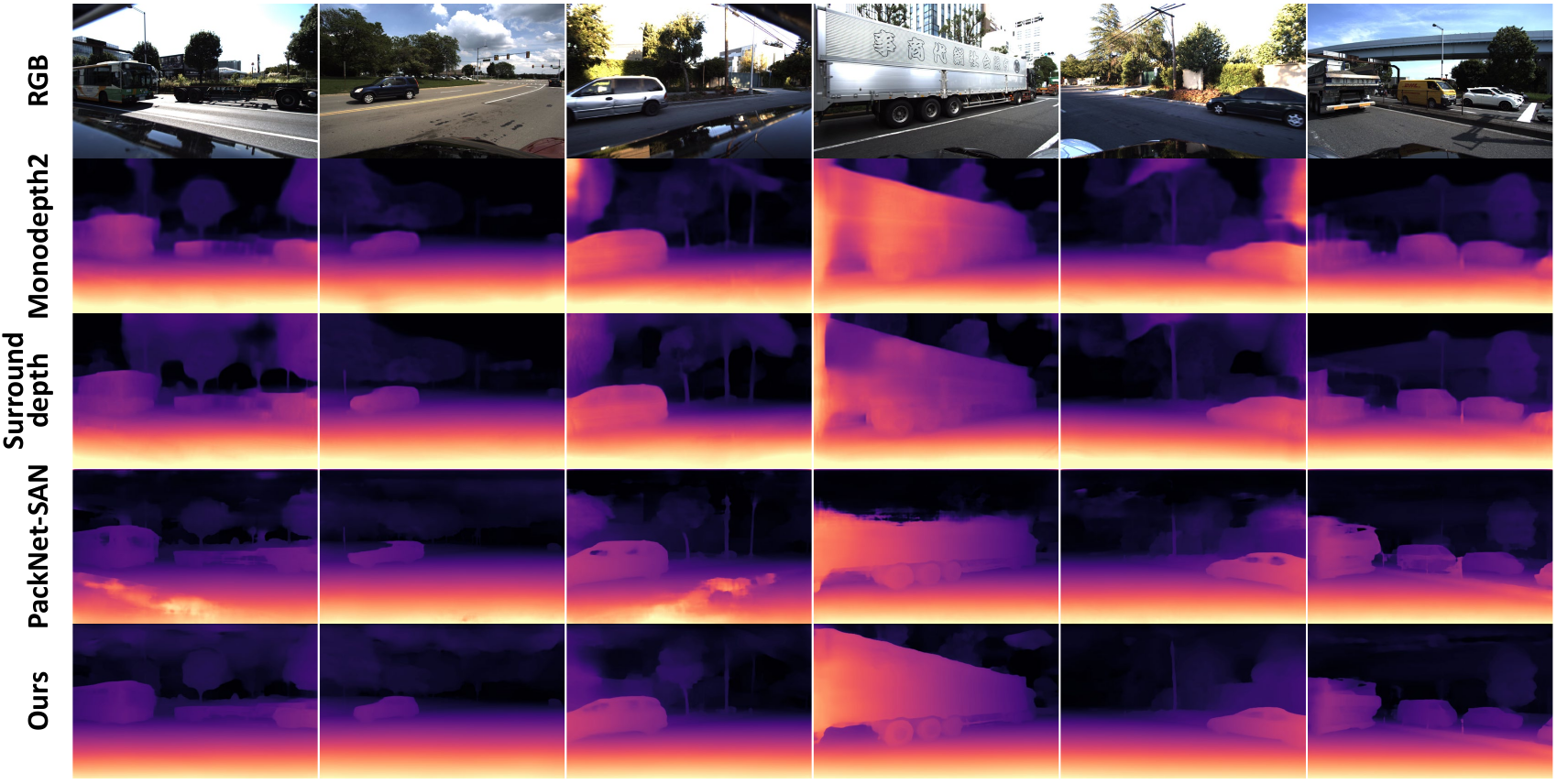}
    \caption{Visual comparisons on \dd{}~\cite{ddad}. We compare our \sxf{} with \sota{} self-supervised~\cite{monodepth2,surround} and supervised~\cite{sans} approaches. Our method shows significant improvements than previous \sota{} supervised PackNet-SAN~\cite{sans}.}
    \label{fig:suppddad}
\end{figure*}

\begin{figure*}[h]
    \centering
    \includegraphics[scale=0.38,trim=0 0 0 0,clip]{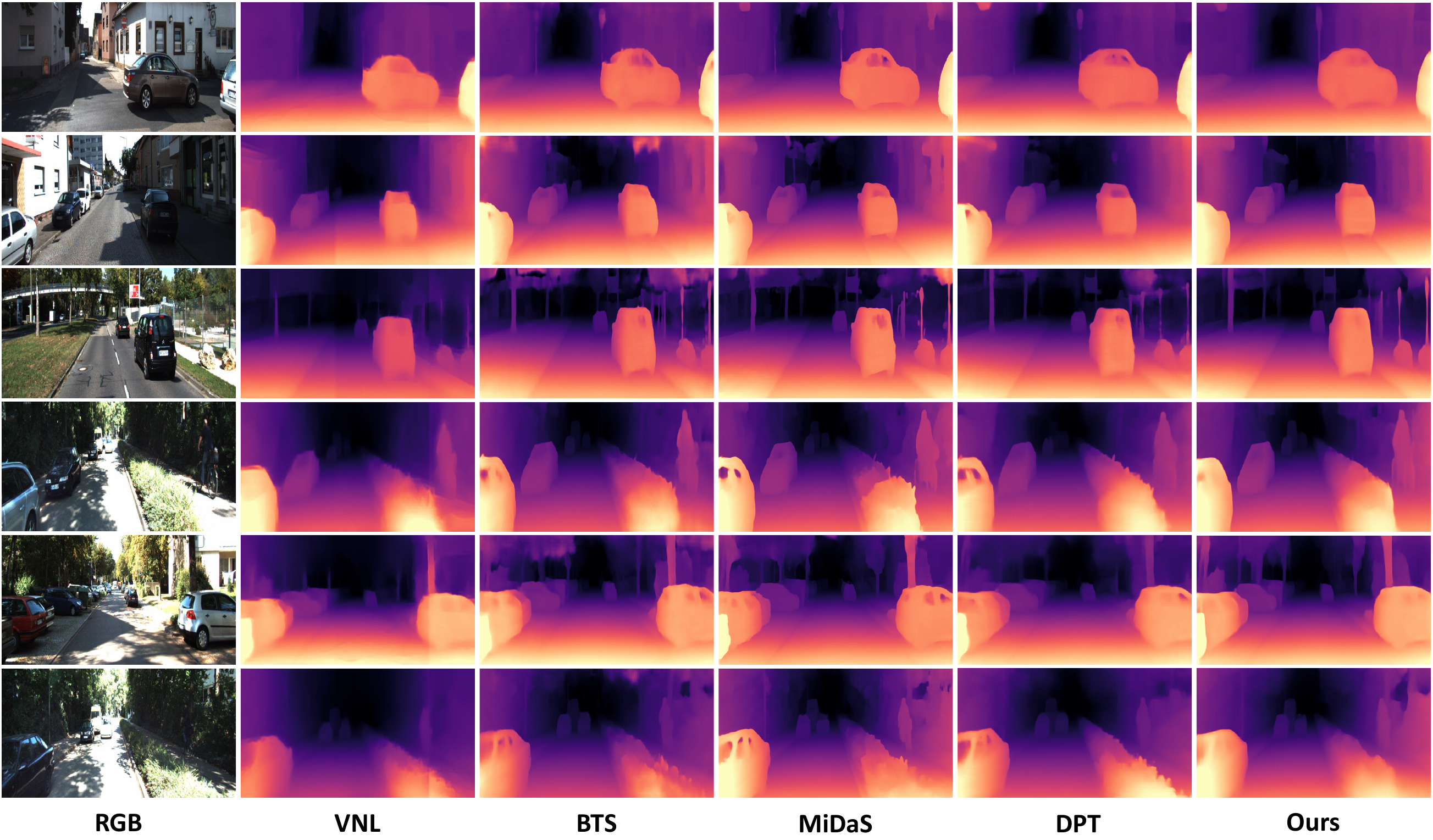}
    \caption{Visual comparisons on \ki{}~\cite{kitti} dataset. We compare \sxf{} with \sota{} supervised~\cite{vnl,bts,midas,dpt} approaches.}
    \label{fig:suppkitti}
\end{figure*}

The visual results on \nus{}~\cite{nus}, \dd{}~\cite{ddad}, and \ki{}~\cite{kitti} datasets are shown in \reffig{}~\ref{fig:suppnus}, \reffig{}~\ref{fig:suppddad}, and \reffig{}~\ref{fig:suppkitti} respectively. In \reffig{}~\ref{fig:suppnus}, we compare our \sxf{} with different depth predictors~\cite{midas,dpt} on \nus{} dataset~\cite{nus} to demonstrate our plug-and-play manner. Comparisons with \sota{} supervised or self-supervised methods on \dd{} dataset~\cite{ddad} are shown in \reffig{}~\ref{fig:suppddad}. The proposed \sxf{} framework shows significant improvements than the previous \sota{} supervised framework PackNet-SAN~\cite{sans} on \dd{} dataset~\cite{ddad}. Visual comparisons with prior arts on KITTI~\cite{kitti} dataset are also shown in \reffig{}~\ref{fig:suppkitti} for sufficient evaluations.


\end{document}